%% file: acl_latex.tex
\documentclass[11pt]{article}

\usepackage[preprint]{acl}
\usepackage{times}
\usepackage{latexsym}

\usepackage[T1]{fontenc}

\usepackage[utf8]{inputenc}

\usepackage{microtype}

\usepackage{inconsolata}
 
\usepackage{graphicx}
\usepackage{booktabs}
\usepackage{amsmath}
\usepackage{amsfonts}
\usepackage{subcaption}  
\usepackage{xcolor}      

\title{Represented but Ignored: A Causal Account of Prosodic Underuse in Audio-Language Models}

\author{Linkai Peng \\
  University of Connecticut \\
  \texttt{linkai.peng@uconn.edu}
  \And
  Baorian Nuchged \\ 
  The University of Texas at Austin \\
  \texttt{baorian@utexas.edu} \\}

\begin{document}
\maketitle
\begin{abstract}
Human speech is richly expressive, with prosody carrying linguistic and emotional information beyond the lexical content. A capable large audio-language model (audio-LLM) should therefore support expressive speech understanding, not only transcribing what was said but also interpreting how it was said. Yet behavioral evaluations alone cannot reveal why a model fails on prosodic input. An error may reflect loss of acoustic information, incorrect internal interpretation, or failure to use a representation that is already available inside the model. We introduce a stage-specific probe ladder for localizing these failure modes in audio-LLMs.\footnote{The code will be released.} Across four understanding-only audio-LLMs, prosodic information is usually preserved in the audio path and decodable in late LLM states. Yet it is only partially expressed in the model's final response. We test the causal status of this latent representation with targeted hidden-state interventions. Every intervention shifts the answer distribution in the predicted direction, and in most model--task cells a single edit at the relevant layer is sufficient to drive the model toward the suppressed prosodic decision, though this recovery is directional rather than a selective restoration of the correct class. Feature-level analysis further suggests that this recoverable signal can be expressed through a small subspace. Some of the highest-attribution features in this analysis align with acoustic cues known to carry prosodic information. Within the matched-content contrasts we test, these results locate the recurring bottleneck not in perceiving prosody but in using it. Models that hear and correctly represent a prosodic cue can still fail to express it in their answers.
\end{abstract}

\input{sections/introduction}
\input{sections/related_work}
\input{sections/experimental_setup}

\input{sections/studies}
\input{sections/conclusion}
\input{sections/limitations}



\bibliography{custom}

\appendix

\input{appendix}

\end{document}

%% file: sections/introduction.tex
\section{Introduction}

\begin{figure}[t]
\centering
\includegraphics[width=\columnwidth]{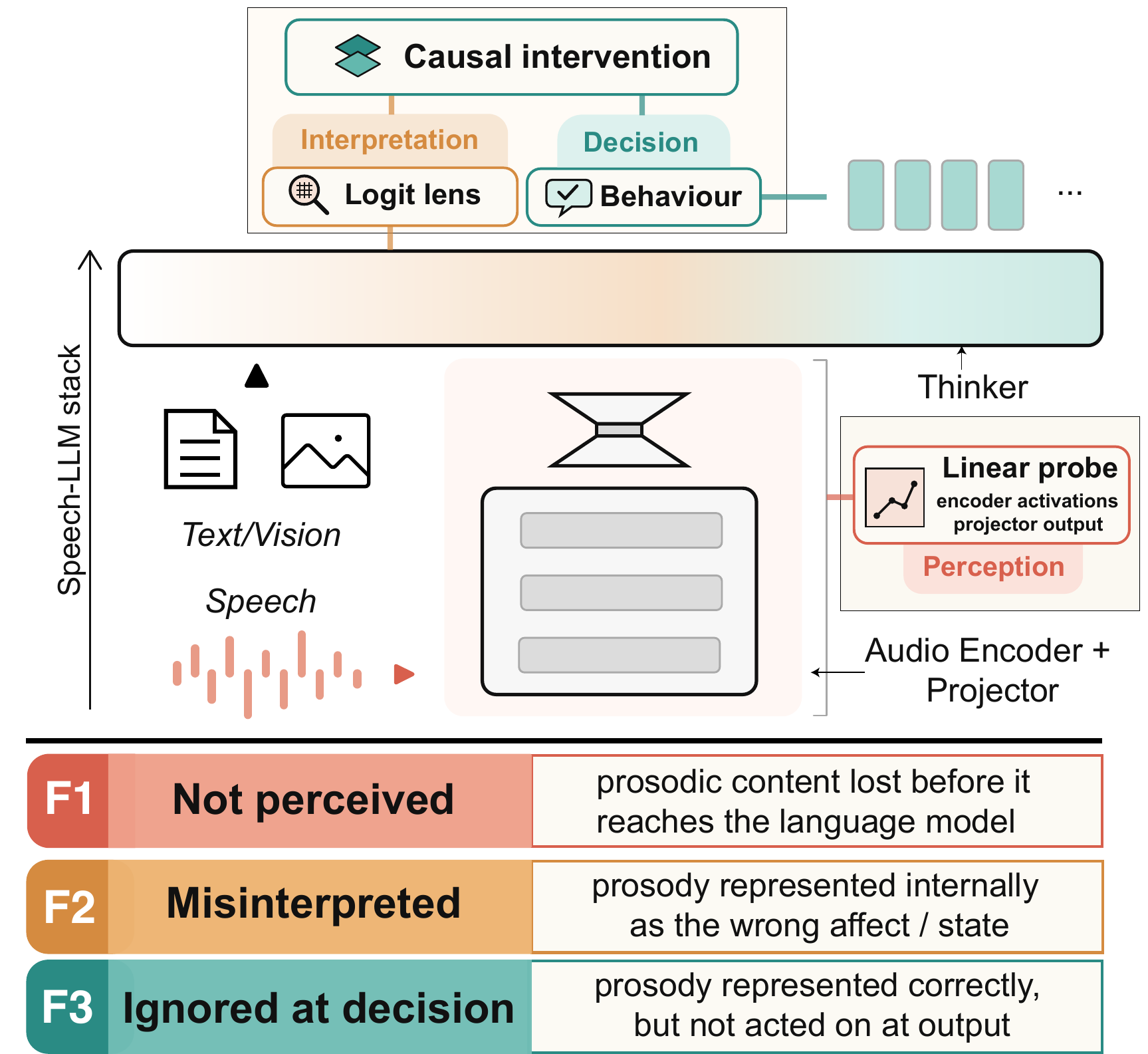}
\caption{Probe ladder for localizing prosody failures in audio-LLMs. The ladder follows prosodic information from the audio pathway, through the LLM representation, to the final answer. The bottom panel maps probe outcomes to three failure modes. In \textbf{F1}, prosody is not perceived; in \textbf{F2}, prosody is internally misinterpreted; and in \textbf{F3}, prosody is represented correctly but underused at the decision stage. The gap between a lens-readable internal signal and behavioral performance defines the \textbf{capability gap}.}
\label{fig:failure-taxonomy}
\vspace{-15px}
\end{figure}

Prosody carries information beyond the lexical content of speech, including both linguistic (intonational) and paralinguistic (emotional) information. Linguistically, intonational contours can distinguish a question from a statement even when the words are identical \citep{pierrehumbert1990meaning,ladd2008intonational}. Paralinguistically, the same sentence can signal different emotional states depending on how it is spoken. These contrasts are carried by acoustic cues such as pitch contour, timing, and intensity \citep{banse1996acoustic,scherer2003vocal,juslin2003communication}. Large audio-language models \citep{qwen25omni,phi4mm,salmonn,audioflamingo,defossez2024moshi} are increasingly expected to use these cues. Prosody-sensitive behavior is now a target of dedicated audio-LLM benchmarks \citep{dynamicsuperb,airbench,sdeval}. Yet their failures are usually evaluated only at the level of the final answer. This makes a behavioral error ambiguous. When a model labels a question as a statement, the failure may lie in the audio pathway, in the internal interpretation of the cue, or in the final decision process. These alternatives are mechanistically distinct and suggest different interventions, but standard benchmarks collapse them into the same outcome.

We formalize this ambiguity as a stage-specific failure taxonomy (Figure~\ref{fig:failure-taxonomy}). In \textbf{F1 perception failures}, the audio pathway does not preserve the relevant prosodic contrast. \textbf{F2 interpretation failures} arise when the model internally represents the wrong prosodic category. Finally, in \textbf{F3 underuse failures}, the correct prosodic category is represented inside the model but is not sufficiently expressed in the final answer. These failure modes are indistinguishable from final-answer accuracy alone, but they imply different bottlenecks and different possible interventions. All three categories are diagnostic abstractions: in real models, a given prosodic task may show evidence for more than one failure source. We therefore assign the dominant failure mode supported by the full probe pattern.

Our probe ladder maps this taxonomy onto stages of the audio-LLM stack. Audio-path probes diagnose perception (F1). For internal interpretation (F2), a layer-wise logit-lens readout \citep{nostalgebraist2020lens,belrose2023eliciting} reads the prosodic category from the residual stream. To diagnose decision use (F3), a behavioral readout anchored between a text-only lexical baseline and a text+cue reference tests whether the represented category is expressed in the final answer. The diagnostic signature for F3 is the \textbf{capability gap}: a decodable internal signal whose behavioral expression falls short of the lexical-cue reference.

We apply this framework to four understanding-only audio-LLMs: Qwen2.5-Omni-7B \citep{qwen25omni}, Phi-4-multimodal-instruct \citep{phi4mm}, Audio-Flamingo-3 \citep{audioflamingo3}, and DeSTA2.5-Audio \citep{desta25audio}. The evaluation uses matched-content contrasts spanning linguistic prosody (question/statement intonation; Q/stmt) and affective prosody (happy/sad, neutral/angry). Encoder and projector probes show that the three Whisper-tower models preserve both contrast types in the representation handed to the LLM. Phi-4's conformer preserves Q/stmt but attenuates matched-content emotion before the projector, which is the single genuine perception (F1) weakness. We next ask whether the LLM misinterprets the preserved cue. A late-layer lens decodes the ground-truth prosodic category in eight of the eleven clean (model $\times$ contrast) cells, so the contrast is internally decodable from the LLM state. However, this internal signal is only partially expressed in behavior. In seven of the eleven cells, audio remains below the text-plus-cue reference despite the decodable code: in six it improves over the text-no-cue baseline but falls short of the reference, and in the limiting case (DeSTA $\times$ IViE) it fails to exceed the baseline at all. The joint pattern (audio-path decodability, correct late-layer lens readout, and incomplete behavioral use) supports F3 underuse as the dominant recurring pattern.

We then ask whether this underused signal is part of the model's own computation or merely an externally readable correlate. At the lens-peak layer ($L^*$),\footnote{The transformer layer at which the logit-lens AUC on the prosodic contrast peaks; formally defined in \S\ref{sec:underuse}.} we apply two interventions. Direction injection adds a class-mean prosodic direction to the answer-position residual stream. Activation patching \citep{vig2020causal,meng2022locating} replaces that state with the corresponding state from a matched donor utterance. Both interventions produce the predicted shifts in answer-token log-odds, and a single linear edit at $L^*$ recovers the gated class on a clean contrast. In addition, sparse autoencoders trained on $L^*$ activations \citep{gao2024scaling}, combined with causal attribution \citep{kramar2024atpstar}, further localize the recoverable signal to a sparse feature subspace. Editing tens of attribution-selected SAE features is sufficient to drive the model toward the targeted prosodic class across the tested mechanism conditions. On the emotion contrasts, top features show acoustic correlates consistent with known vocal-arousal cues.

This paper makes four contributions. First, we introduce a stage-specific taxonomy and probe ladder for distinguishing prosody failures that are indistinguishable from final-answer accuracy alone. Second, we show that, across four understanding-only audio-LLMs, F3 underuse is the dominant recurring pattern in the matched-content conditions we test. Prosody can be internally recoverable while remaining only partially expressed in behavior. Third, we provide causal evidence that the late-layer prosodic signal is causally controllable: a single linear edit at $L^*$ is sufficient to drive the model toward the gated class, though the effect is a directional bias rather than a selective restoration of the correct decision. Fourth, we localize the recoverable signal to a small SAE feature subspace, with top-attribution features on the emotion contrasts showing acoustic correlates consistent with known vocal-arousal cues. Together, these results suggest that, within the matched conditions we test, a recurring failure mode is not a failure to hear prosody. Instead, it is a failure to use an available internal representation.

%% file: sections/related_work.tex
\section{Related Work}

\paragraph{Audio-LLMs and prosody-sensitive evaluation.}
Audio-language models route speech through an encoder and projector into a text LLM \citep{zhang2023speechgpt,qwen2audio,salmonn,audioflamingo,qwen25omni,phi4mm}. Benchmarks such as Dynamic-SUPERB, AIR-Bench, and SD-Eval probe prosody-sensitive behavior including question/statement classification, sarcasm, and emotion recognition \citep{dynamicsuperb,airbench,sdeval}. These benchmarks score only the final answer, so a wrong label cannot be attributed to failed perception, misinterpretation, or unused internal representation. Our F1/F2/F3 taxonomy assigns each failure to a stage.

\paragraph{Representation probing on speech encoders.}
Probing studies show that prosodic categories are linearly recoverable from self-supervised and ASR-trained speech encoders \citep{hubert,wavlm,whisper}. Decodability rises with encoder scale \citep{pasad2021layerwise,deseyssel2022probing,wagner2023dawn,wang2021finetuned,emotion2vec}. This is the linear-probe baseline against which the audio-tower and projector probes of \S\ref{sec:study-1} are calibrated (App.~\ref{app:probes}). What this work does not address is what happens to prosody once it crosses into the LLM. That gap motivates the lens, intervention, and feature-attribution analyses in \S\ref{sec:underuse}--\S\ref{sec:sfc}.

\paragraph{Mechanistic interpretability for language models.}
Mechanistic interpretability of language models has produced three relevant tools, all developed on text-only LLMs. These are logit-lens readouts \citep{nostalgebraist2020lens,belrose2023eliciting}, hidden-state interventions like direction injection and activation patching \citep{vig2020causal,geiger2021causal,meng2022locating,subramani2022extracting,wang2023interpretability}, and sparse-feature attribution \citep{elhage2022superposition,cunningham2024sparse,templeton2024scaling,marks2025sparse}. From this toolkit we adopt the TopK sparse autoencoder of \citet{gao2024scaling} and the AtP\textsuperscript{*} attribution method of \citet{kramar2024atpstar}. We apply them to a failure mode with no text-only analog, where the audio path, the LLM, and the output decision must be tracked jointly.

Our contribution is to connect these levels: we diagnose where prosody fails in the audio-LLM stack, then test whether the internally represented signal is causally controllable and recoverable.

%% file: sections/experimental_setup.tex
\section{Experimental Setup}
\label{sec:setup}

This section defines the models and corpora. Study-specific procedures are introduced in the corresponding study sections.

\subsection{Models}
\label{sec:setup-models}

We evaluate four understanding-only audio-LLMs. \textbf{Qwen2.5-Omni-7B} \citep{qwen25omni} and \textbf{Audio-Flamingo-3} \citep{audioflamingo3} use a Whisper-style audio tower and a 28-layer language model. \textbf{DeSTA2.5-Audio} \citep{desta25audio} pairs a frozen Whisper-large-v3 encoder with a 32-layer Llama-3.1-8B language model through a Q-Former adapter. \textbf{Phi-4-multimodal-instruct} (Phi-4-MM) \citep{phi4mm} uses a conformer audio tower \citep{gulati2020conformer} and a 32-layer language model. The mechanistic analyses are restricted to models for which answer-position hidden states can be extracted and intervened on, enabling logit-lens readout, activation patching, and SAE-based feature interventions. Audio-path probes in \S\ref{sec:study-1} additionally use off-the-shelf speech encoders as calibration references for what linear classifiers can decode from standard speech representations. Checkpoints and architecture details are in Appendix~\ref{app:models}.

\subsection{Datasets}
\label{sec:setup-data}
 
We use corpora spanning intonation prosody and emotional prosody (Table~\ref{tab:datasets}). Our primary evidence comes from \textbf{matched-content} settings, where lexical content is controlled so the target distinction is carried by prosody alone. These settings use same-speaker utterances with matched or near-matched lexical content but opposite prosodic labels. \textbf{IViE} \citep{grabe2001ivie} (question/statement intonation), \textbf{CREMA-D} \citep{cao2014cremad} (emotion), and \textbf{VESUS} \citep{sager2019vesus} (emotion) provide these matched-content contrasts. A further reason for this selection is training contamination. Audio-Flamingo-3 and DeSTA2.5-Audio include widely used emotion corpora in their training data, which narrows the set of corpora usable for a clean held-out evaluation. Additional corpora (JL-Corpus \citep{james2018jlcorpus} and ESD-English \citep{zhou2022esd}) serve for audio-path calibration and supplementary analyses, with in-training (model $\times$ corpus) cells masked. Per-class counts, filtering rules, and per-study assignments are given in Appendix~\ref{app:setup}.

\begin{table}[t]
\small
\centering
\setlength{\tabcolsep}{4.5pt}
\begin{tabular}{llcc}
\toprule
Corpus & Contrast & $n$ & Patchable pairs \\
\midrule
IViE         & Question/statement & 430      & \checkmark \\
CREMA-D      & 4-class emotion    & 4{,}348  & \checkmark \\
VESUS        & 4-class emotion    & 10{,}073 & \checkmark \\
\bottomrule
\end{tabular}
\caption{Corpora used as the primary matched-content contrasts.}
\label{tab:datasets}
\vspace{-10px}
\end{table}

\subsection{Prompts}
\label{sec:setup-prompts}

The behavioral readout in \S\ref{sec:underuse} queries the LLM with a prompt that names the prosodic class of the audio clip. Intervention readouts in \S\ref{sec:causal}--\S\ref{sec:sfc} use the same prompting. Each architecture's official chat template is used unmodified. Prompt format varies by contrast type: forced-choice for binary contrasts, mixed direct-question and multiple-choice for the 4-class emotion evaluation. To rule out brittleness to any single phrasing, each clip is scored under $N$ paraphrased prompts; we report the per-clip \textbf{majority vote} as the verdict for that contrast. Audio-path probes in \S\ref{sec:study-1} do not involve prompting. Per-condition $N$, the full prompt banks, system-prompt strings, pole-token lists, and output-parsing rules are in Appendix~\ref{app:prompts}.

%% file: sections/studies.tex
\section{Studies}
\label{sec:studies}
\subsection{Is prosody perceived in the audio path?}
\label{sec:study-1}

We first ask whether prosodic information is available to the language model. For each of the four audio-LLMs we train layer-wise linear probes on the frozen audio tower. Weighted accuracy (WA)\footnote{Since we balance the dataset classes, WA equals raw accuracy.} is read against a random-initialized baseline run through the same pipeline. Because that baseline differs across audio front ends, we diagnose F1 from the lift over the baseline rather than absolute WA. Probe validity is established by replicating known decodability results on six off-the-shelf encoders (Appendix~\ref{app:probes}).

\begin{figure}[t]
\centering
\includegraphics[width=\columnwidth]{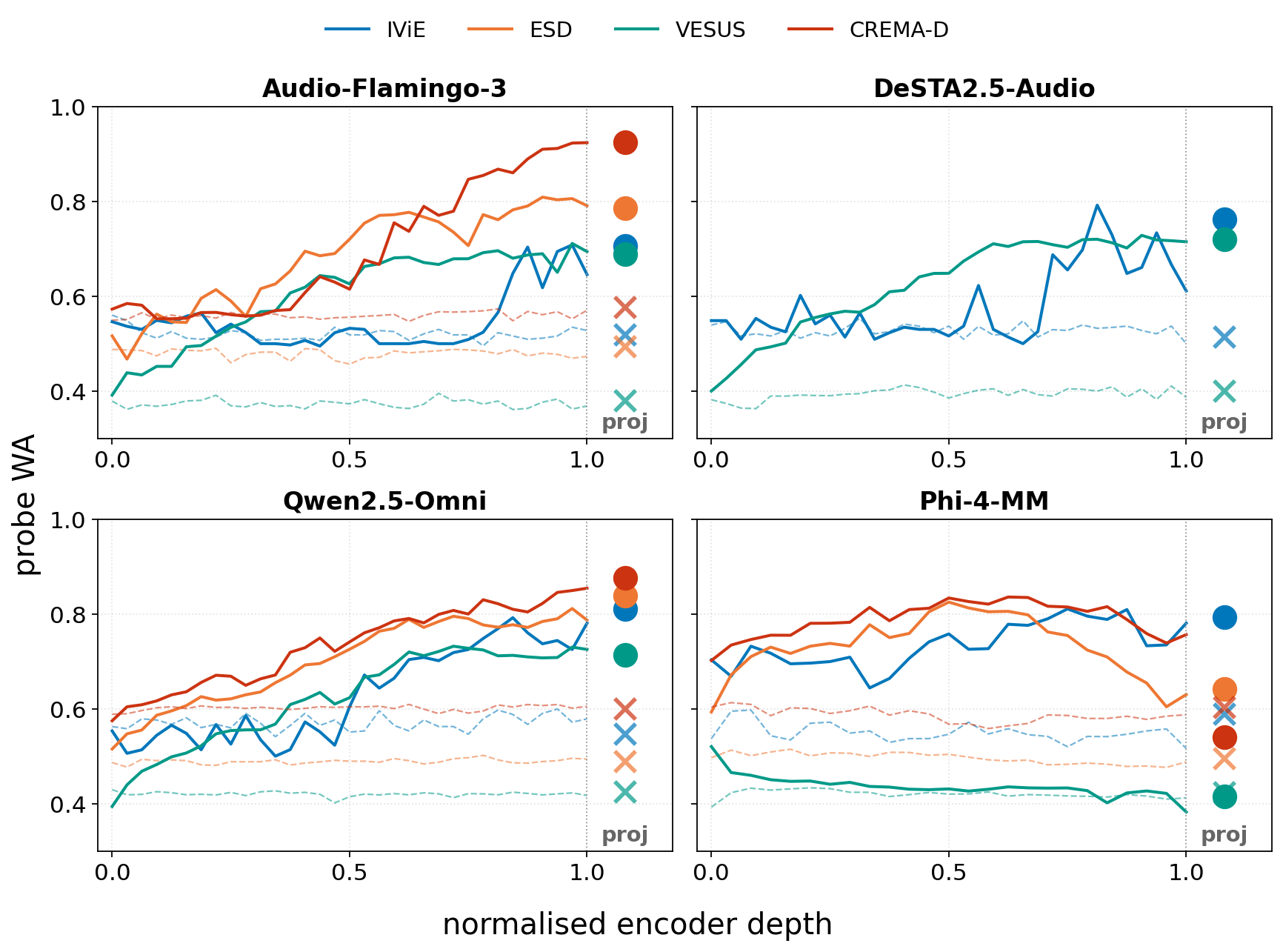}
\caption{Audio-path probes for the four audio-LLMs, one panel per model. Each panel shows layer-wise probe WA for IViE Q/stmt and three emotion corpora. The $x$-axis is normalized encoder depth (0 = input, 1.0 = final encoder block). Markers at PROJ are the post-projector representation passed to the LLM. Solid lines are the trained encoder, dotted the random-initialized baseline (same pipeline). DeSTA2.5's CREMA-D and ESD cells lie in its training data and are masked. Exact per-cell numbers, cross-validation folds, and held-out / training-overlap routing are in Appendix~\ref{app:probes}.}
\label{fig:audio-path-probes}
\vspace{-15px}
\end{figure}

Across the four corpora (Figure~\ref{fig:audio-path-probes}), the three Whisper-tower models---Qwen2.5-Omni (Qwen), Audio-Flamingo-3 (AF3), and DeSTA2.5 (DeSTA)---preserve a strong prosodic signal through the audio path. At the projector handed to the LLM, every emotion cell sits well above its random-initialized baseline. WA is 0.79--0.92 on CREMA-D and ESD and 0.69--0.72 on VESUS (lift $+0.28$ to $+0.35$). IViE Q/stmt reaches WA 0.71--0.81 (lift $+0.19$ to $+0.26$). DeSTA2.5 freezes a Whisper-large-v3 encoder, so its layer curves track vanilla Whisper directly; its VESUS projector (0.72) falls in the same band as the other two models. For all three, both intonational and emotional prosody remain decodable in the representation passed to the LLM, and a pure F1 (not-perceived) account is disfavored on every tested contrast.
 
Phi-4-MM is contrast-specific. On IViE Q/stmt the conformer carries the signal through the last encoder layer and into the projector (WA 0.79, lift $+0.21$). F1 is thus also disfavored for Phi-4 on intonation. Across CREMA-D and ESD, the conformer represents emotion mid-stack (peak WA $\approx 0.84$ at depth 0.5--0.6, consistent with \citealp{morais2024conformer}). On the VESUS corpus, the conformer barely encodes emotion at any depth (peak WA $\le 0.52$) and sits even under the random baseline. This is far below the $\approx 0.72$ the Whisper-tower models reach on the same clips. For Phi-4, matched-content emotion is therefore F1-limited: the prosodic signal is at most partially represented in the conformer and is not preserved in the representation handed to the LLM.

We therefore treat the F1 verdict as architecture- and contrast-specific. Pure F1 is disfavored for Qwen, AF3, and DeSTA across the tested intonational and emotional contrasts and for Phi-4 on IViE Q/stmt. The single genuine F1 weakness is Phi-4 on matched-content emotion.
 
\begin{figure*}[t]
\centering
\includegraphics[width=\textwidth]{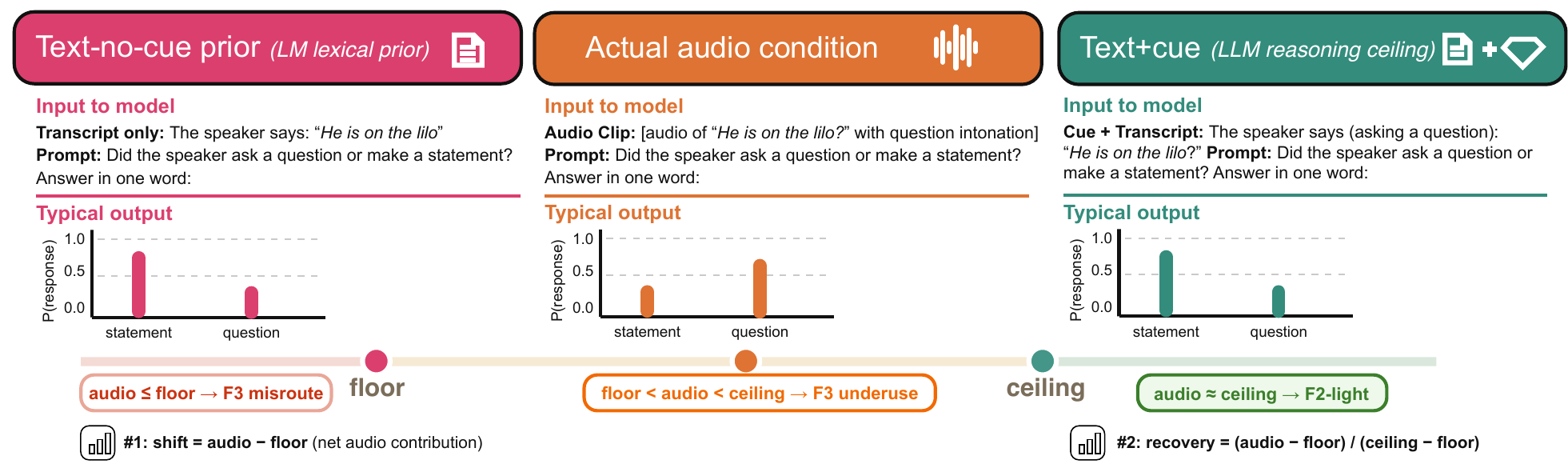}
\caption{Three-condition ladder schematic. The audio's position on the recall axis sits between the text-no-cue baseline (LM lexical prior) and the text+cue reference. The reference is the model's response when the prosodic state is given explicitly in text. The ladder measures audio's behavioral contribution; the final F-mode diagnosis combines this ladder position with the internal lens readout.}
\label{fig:ladder-schematic}
\vspace{-10px}
\end{figure*}

\subsection{Do audio-LLMs use represented prosody?}
\label{sec:underuse}

Representation in the audio path does not entail that the LLM acts on prosody. A prosody-sensitive behavioral error is compatible with two downstream failures. The model may internally represent the wrong prosodic category (F2). Or it may represent the correct category but fail to express it in its answer (F3). These modes are indistinguishable from final-answer accuracy alone. We therefore evaluate each clip with two coupled readouts under matched prompt conditions. An \emph{internal} readout asks whether the prosodic category is decodable from the LLM's hidden states. A \emph{behavioral} readout estimates how much audio contributes to the generated answer relative to the model's text-modality prior.

\paragraph{Internal readout.}
The internal readout combines two complementary measures of the answer-position residual stream. One is a logit \textbf{lens} \citep{nostalgebraist2020lens,belrose2023eliciting}. At each layer, we apply the model's unembedding matrix and compute the relative log-odds of the relevant prosodic-pole token sets (such as ``question'' versus ``statement''). For binary contrasts, we report binary AUC; for 4-class emotion, one-vs-rest AUC per class, macro-averaged. The layer with the highest AUC, $L^*$, summarizes where the residual stream most strongly carries the prosodic category. This lens asks whether the category is decodable through the model's \emph{own} readout. To quantify \emph{how much} usable information the state carries, we additionally report predictive \textbf{V-information} \citep{xu2020vinformation}, $I_{\mathcal V}(h_{L^*}\!\to Y)$. It is the reduction in label entropy achievable by a bounded predictor family $\mathcal V$. Here $\mathcal V$ comprises L2-regularized linear softmax probes on the $L^*$ state. The family matches the linearity of the model's own unembedding, so the measured bits sit in the model's access basis. We report $I_{\mathcal V}$ in bits against the label-entropy ceiling (1.0 for Q/stmt, $\approx$2.0 for balanced 4-class emotion). It is estimated on held-out clips with bootstrap 95\% confidence intervals and validated with a shuffled-label control \citep{hewitt2019control} that yields $I_{\mathcal V}\approx 0$. Unlike rank-based AUC, it cannot be inflated by thin threshold-free separability (Appendix~\ref{app:vinfo}).

\paragraph{Behavioral readout.}
The behavioral readout isolates audio's net prosodic contribution from the LLM's lexical prior on the transcript. For binary contrasts, we score recall on the prosodic positive class. On 4-class emotion, we score macro-averaged per-class recall.\footnote{Recall rather than precision because the F3 failure of interest is a false negative on the positive prosodic pole.} Raw recall alone is insufficient, because it conflates audio use with the LM's prior over the transcript. We therefore use a \emph{three-condition ladder} (Figure~\ref{fig:ladder-schematic}). It adds a text-no-cue \emph{baseline}, using the transcript only, and a text+cue \emph{reference}, using the transcript plus an explicit prosodic-state cue. From the three recall values, we compute two summaries. The first is $\text{shift}=\text{audio}-\text{baseline}$, the net behavioral contribution of audio over the text prior. Second, $\text{\% recovery}=\text{shift}/(\text{reference}-\text{baseline})$ gives the fraction of the available baseline-to-reference range recovered by audio (Appendix~\ref{app:prompts}).

The ladder is interpreted jointly with the lens. When $\text{audio}\le\text{baseline}$, audio fails to add useful information over the lexical prior. Under low lens AUC this is a misroute or no-net-audio-contribution case. With high lens AUC it is instead the limiting case of F3 (\emph{complete} underuse: represented but not used at all). If instead $\text{baseline}<\text{audio}<\text{reference}$ and lens AUC is high (operationally, peak AUC $\geq 0.75$), the signature is F3 underuse. Here the correct category is represented but not fully expressed in behavior. The same ladder pattern with low lens AUC (peak AUC $\leq 0.65$) is better interpreted as partial F2, because the category is not cleanly represented in the LLM state. Finally, $\text{audio}\approx\text{reference}$ indicates no behavioral underuse relative to the model's text-cue reference.\footnote{A sub-1.0 reference is not itself evidence of underuse: the reference is a practical upper bound set by the model's response to an explicit text cue.} Baseline and reference protocol implementation is described in Appendix~\ref{app:text-control}.
 
\begin{figure}[t]
\centering
\includegraphics[width=\columnwidth]{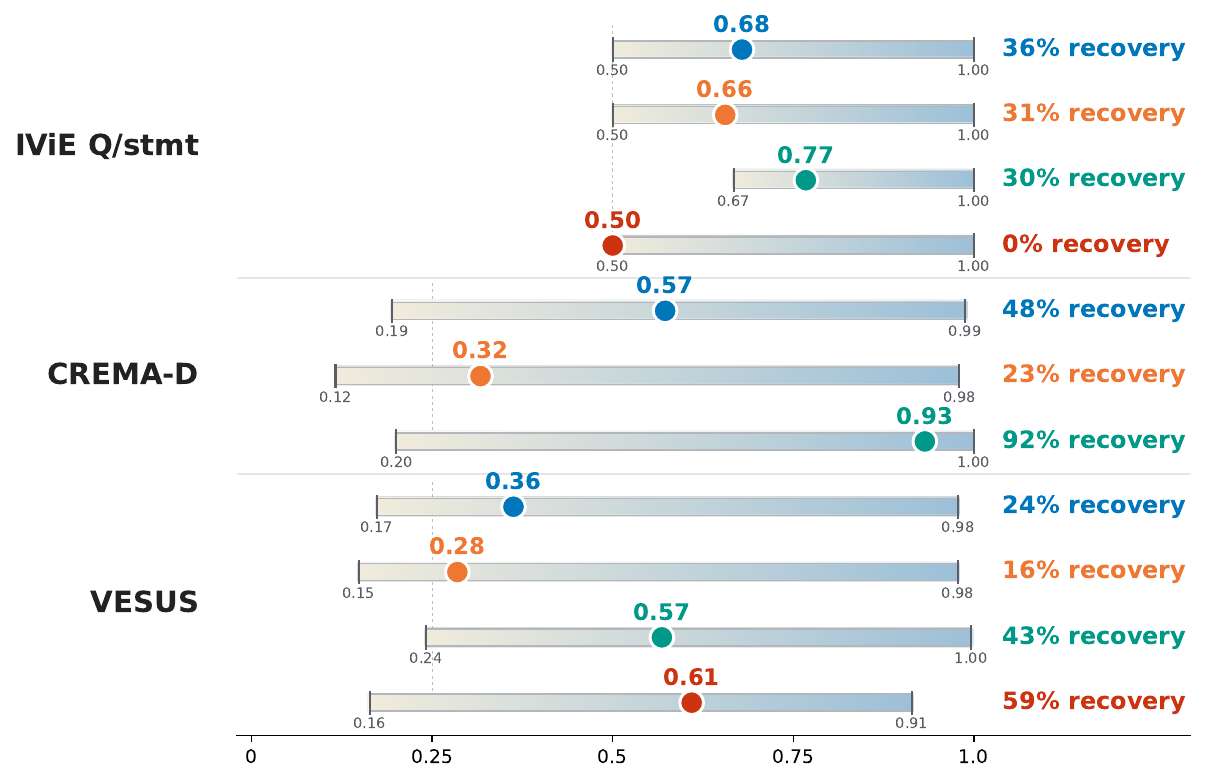}
\caption{Behavioral ladders for the three main cells. Color identifies the model. Top to bottom within each block: \textbf{\textcolor[HTML]{0077BB}{Qwen2.5-Omni}}, \textbf{\textcolor[HTML]{EE7733}{Phi-4-MM}}, \textbf{\textcolor[HTML]{009988}{Audio-Flamingo-3}}, \textbf{\textcolor[HTML]{CC3311}{DeSTA2.5}}. Each bar spans the text-no-cue \emph{baseline} to the text+cue \emph{reference}; the dot marks the \emph{audio} condition and the right tag its \% recovery of that range. DeSTA $\times$ CREMA-D is masked (in-training). The supplementary corpus (ESD) is in Appendix~\ref{app:underuse-supplement}.}
\label{fig:results-ladder}
\vspace{-10px}
\end{figure}

\input{tab_lens_ladder_verdicts}

\paragraph{Results.}
Table~\ref{tab:lens-ladder-verdicts} reports both readouts for the eleven clean (model $\times$ contrast) cells, with DeSTA $\times$ CREMA-D masked as in-training. Layer-wise lens curves are in Appendix~\ref{app:lens-curves}. The late stack carries a strong prosodic code in eight of the eleven cells. Lens AUC is 0.81--1.00, with $I_{\mathcal V}$ of 1.49--1.75 bits of the 2.0-bit label on the strong emotion cells. Three weak codes remain: Phi-4's emotion cells (AUC 0.56--0.61) and Qwen $\times$ VESUS (0.63). The two measures agree on emotion but diverge on intonation: the IViE cells reach AUC 0.81--0.83 yet only 0.23--0.41 bits of $I_{\mathcal V}$ (of a 1.0-bit ceiling) with wide CIs, indicating that the question/statement code is rank-separable but carries less bounded-probe information than the high AUC alone implies. The IViE F3 verdicts therefore rest on the AUC gate, and intonation decodability should be read as weaker than the emotion cases.

Behavior expresses only part of this information (Figure~\ref{fig:results-ladder}). On IViE, Qwen, Phi-4, and AF3 recover 30--36\% of the available range, while DeSTA recovers 0\%. DeSTA answers ``statement'' on every clip unless the cue is written out. On emotion, AF3 nearly closes the gap on acted CREMA-D (92\%), yet on VESUS even the best models leave roughly 40--60\% of the range unused (AF3 43\%, DeSTA 59\%). Qwen recovers 48\% on CREMA-D but pairs a weak code with weak behavior on VESUS (partial F2), and Phi-4's emotion cells inherit the \S\ref{sec:study-1} F1 verdict.

In sum, seven of the eleven cells are F3 underuse, and the exceptions are systematic rather than random. Underuse is sharpest where representation is strongest (DeSTA $\times$ IViE) and persists on held-out VESUS for every model that represents the contrast. AF3 carries 1.55 bits there while expressing 0.57 macro recall. ESD-English replicates the pattern (Appendix~\ref{app:underuse-supplement}).

\subsection{Can a single intervention recover the underused signal?}
\label{sec:causal}

\begin{figure}[t]
\centering
\includegraphics[width=\columnwidth]{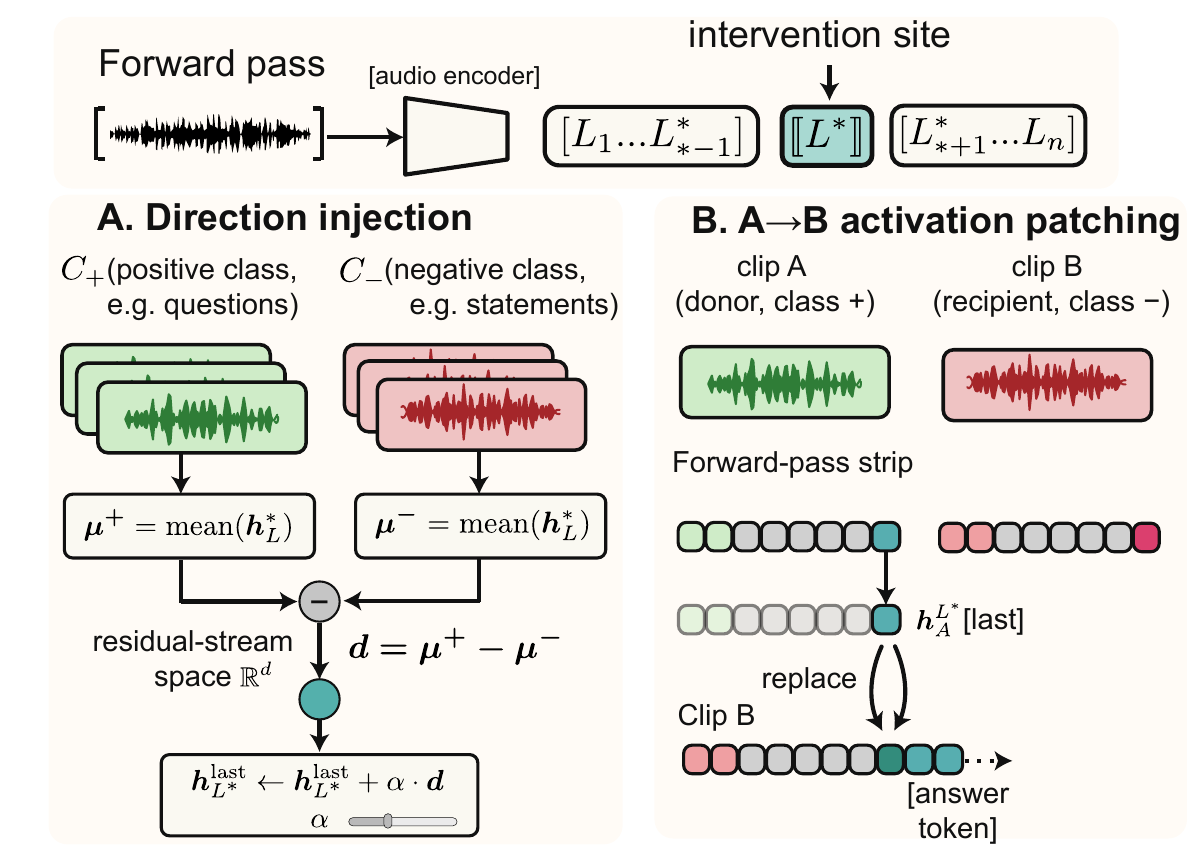}
\caption{Two interventions at the lens-peak layer $L^*$. \textbf{(A) Direction injection.} A class-mean direction $d=\mu^+-\mu^-$ is computed from positive- and negative-class $L^*$ states. The answer-position residual stream is then edited as $h_{L^*}^{\mathrm{last}} \leftarrow h_{L^*}^{\mathrm{last}}+\alpha d$. \textbf{(B) Activation patching.} For a matched pair, we run a donor clip $A$ to $L^*$. We then insert its answer-position state into the recipient clip $B$ at the same layer and position. The rest of $B$'s forward pass is unchanged. Both interventions test whether changing the $L^*$ answer-position state changes the model's answer-token distribution.}
\label{fig:intervention-schematic}
\vspace{-15px}
\end{figure}

\paragraph{Decodability alone does not establish causality.}
\S\ref{sec:underuse} showed the F3 signature. Prosodic category information is decodable at a late layer $L^*$, but the model's behavioral answer expresses only part of that information. This decodability gap is suggestive, but not yet causal. A signal at $L^*$ could be an externally readable correlate that downstream layers do not use. Alternatively, it could lie on a path that influences the answer but is underweighted in the unedited forward pass. We test these possibilities with two single-site interventions at $L^*$ (Figure~\ref{fig:intervention-schematic}). First, \textbf{direction injection} moves the answer-position state along a class-mean prosodic direction. Second, \textbf{activation patching} replaces that state with a naturally occurring state from a matched donor utterance. From this section onward, emotion is split into two binary contrasts (happy/sad and neutral/angry). Together with IViE Q/stmt, this gives five binary contrasts per architecture.

\paragraph{Direction injection shows causal coupling.}
For each binary contrast, we estimate a class-mean direction in the $L^*$ residual stream, $d=\mu^+-\mu^-$. Here $\mu^+$ and $\mu^-$ are the mean answer-position states for the positive and negative classes, computed on clips disjoint from the intervention set. At inference time, we run the model normally up to $L^*$, add $\alpha d$ only to the answer-position state, and continue the forward pass unchanged. All model parameters, other token positions, and other layers are left fixed. We then measure the change in prosodic answer-token log-odds as $\alpha$ varies. Full details of the log-odds metric, the $\alpha$ grid, prompt-level results, and confidence intervals are given in Appendix~\ref{app:causal-injection}. Table~\ref{tab:di-slopes} reports the direction-injection slopes across the eighteen clean cells. All eighteen slopes are positive. Moving the $L^*$ answer-position state toward the positive-class mean therefore moves the model's output distribution toward the corresponding positive answer token. This does not by itself prove that the unedited forward pass fully uses this state. It does, however, strongly disfavor a purely epiphenomenal interpretation of the lens signal. The signal is not merely readable by an external probe, because changing it systematically changes the model's own answer-token distribution.

\input{tab_di_slopes_4model}

\paragraph{Activation patching validates the effect with natural states.}
Direction injection uses an analytic mean-difference vector. To test whether naturally occurring $L^*$ states produce the same directional effect, we next use activation patching. For each matched pair, we replace the recipient clip's answer-position state at $L^*$ with the corresponding state from an opposite-class donor clip, then continue the recipient's forward pass. The diagnostic prediction is bidirectional: a negative-to-positive patch should shift the output toward the positive pole, while a positive-to-negative patch should shift it toward the negative pole.

Table~\ref{tab:patching} reports the patching shifts on VESUS for all four models. All sixteen shifts take the predicted sign, and effects are approximately antisymmetric within each cell. This strengthens the causal coupling result because it does not depend only on a synthetic class-mean direction. Naturally occurring opposite-class states at $L^*$ also move the answer distribution in the predicted direction. Magnitude tracks representation strength: AF3 and DeSTA show the largest effects ($|\Delta|$ 3.3--5.2), Phi-4 the smallest ($\le$0.6), consistent with the \S\ref{sec:underuse} gradient. The CREMA-D and IViE patching cells show the same signed, antisymmetric pattern (Appendix~\ref{app:causal-patching-cells}).

\input{tab_patching_vesus}

\paragraph{A single $L^*$ intervention recovers the gated class.}
The log-odds shifts above indicate that the $L^*$ state can move the answer distribution, but not yet that the decoded answer changes. We therefore sweep $\alpha$ and classify each clip by the sign of its answer-token log-odds. On the Qwen $\times$ CREMA-D neutral/angry cell, angry recall rises from 0.366 at $\alpha=0$ to 1.000 by $\alpha=2$. Across the grid, recall saturates in fifteen of the eighteen clean cells within the tested $\alpha$ range (Table~\ref{tab:di-recall-sweep}). The recovery is directional rather than symmetric: pushed far enough, some negative-class clips also cross the decision boundary. Appendix~\ref{app:causal-sweep} reports the full sweeps.

\subsection{What sparse features support the recoverable signal?}
\label{sec:sfc}
We next ask whether the same recoverability can be expressed in a sparse feature basis. This section is therefore a sparse-space replication of \S\ref{sec:causal}. Instead of moving the full residual stream along a dense class-mean direction, we intervene only on a small set of attribution-selected SAE features at $L^*$. Architecture-specific TopK sparse autoencoders (keeping only the $k$ largest latent activations per token) are trained on $L^*$ activations sampled across a multi-corpus speech pool. Features are then ranked by AtP$^*$ attribution to the prosodic log-odds. Let $S_{0.95}$ denote the smallest feature set covering 95\% of absolute attribution. Clamp formulas (\S\ref{app:sfc-clamp}) and SAE training (\S\ref{app:sfc-training}) are detailed in Appendix~\ref{app:sfc}. So are AtP$^*$ attribution and $S_{0.95}$ selection (\S\ref{app:sfc-attribution}).

\paragraph{Sparse-space intervention reproduces dense recovery.}
Attribution is concentrated. Across the eighteen clean (model $\times$ contrast) cells, $S_{0.95}$ contains only 33--102 features, never more than 0.5\% of the SAE dictionary (Figure~\ref{fig:sfc-attribution}). We then sweep an additive clamp on only the $S_{0.95}$ coordinates of the answer-position SAE code toward their positive-class activation pattern. This sparse intervention recovers the positive prosodic class on 17 of 18 clean cells (Table~\ref{tab:sfc-sufficiency}). The single exception is Phi-4 $\times$ VESUS neutral/angry, the weakest model on the hardest held-out contrast. Thus the dense recoverability result from \S\ref{sec:causal} is not limited to an opaque residual-stream direction: the same gated answer can be exposed through a small attribution-selected subspace.

\begin{figure*}[t]
\centering
\includegraphics[width=\textwidth]{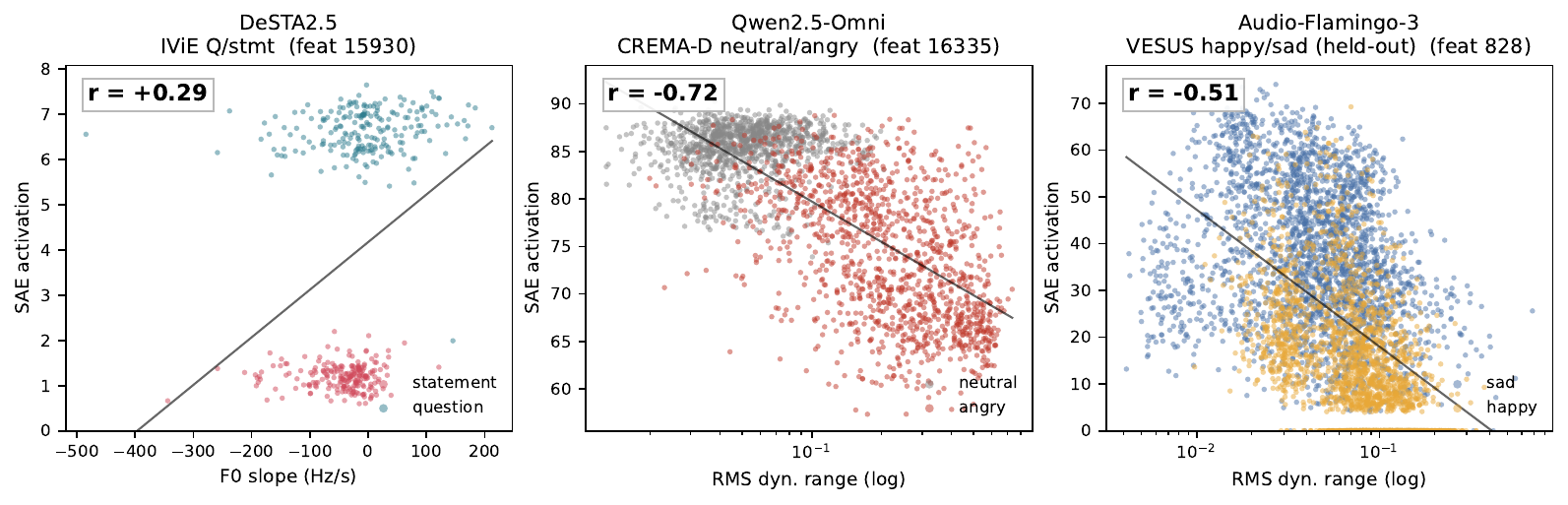}
\caption{Acoustic correlates in sparse features. One example cell from each of three models, chosen to cover the three datasets. The cells are DeSTA on IViE question/statement, Qwen on CREMA-D neutral/angry, and AF3 on held-out VESUS happy/sad. Each panel shows the strongest class-appropriately-aligned feature among the cell's top-3 attribution features. It is plotted against the expected acoustic cue (F0 slope for intonation, RMS-energy variation for emotion). Dots are individual clips; the line is an OLS fit; $r$ is the Pearson correlation across the cell's clips. Alignment is strong for the emotion cells and weak for intonation (see text); per-model, per-cell grids are in Appendix~\ref{app:sfc-mechanism-grids}.}
\label{fig:sae-mechanism-grid}
\vspace{-15px}
\end{figure*}

\begin{table}[t]
\centering
\scriptsize
\setlength{\tabcolsep}{4pt}
\begin{tabular}{lcccc}
\toprule
Cell & Qwen & Phi-4 & AF3 & DeSTA \\
\midrule
IViE Q/stmt & 0.46$\to$\textbf{1.00} & 0.01$\to$\textbf{1.00} & 0.63$\to$\textbf{1.00} & 0.03$\to$\textbf{1.00} \\
CREMA-D h/s & 0.87$\to$\textbf{1.00} & 0.89$\to$\textbf{1.00} & 0.98$\to$\textbf{1.00} & \emph{masked} \\
CREMA-D n/a & 0.17$\to$\textbf{1.00} & 0.33$\to$\textbf{1.00} & 0.97$\to$\textbf{1.00} & \emph{masked} \\
VESUS h/s & 0.39$\to$\textbf{1.00} & 0.96$\to$\textbf{0.99} & 0.68$\to$\textbf{1.00} & 0.55$\to$\textbf{1.00} \\
VESUS n/a & 0.12$\to$\textbf{1.00} & 0.09$\to$\textbf{0.78} & 0.57$\to$\textbf{1.00} & 0.59$\to$\textbf{1.00} \\
\bottomrule
\end{tabular}
\caption{Feature-level sufficiency across four models. Each entry is positive-class recall at $\alpha{=}0$ (SAE reconstruction, no clamp) $\to$ peak under the $S_{0.95}$ additive-diff clamp. DeSTA $\times$ CREMA-D is masked as in-training. Full $\alpha$-sweeps in Table~\ref{tab:sfc-recovery}.}
\label{tab:sfc-sufficiency}
\vspace{-15px}
\end{table}
 
\paragraph{Sparse features track acoustic cues most strongly for vocal arousal.}
We next ask whether the selected features are interpretable in acoustic terms. For each cell, we correlate the top-3 attribution features with six per-clip acoustic descriptors covering F0, duration, and RMS energy (Appendices~\ref{app:sfc-acoustic}--\ref{app:sfc-desc}). These descriptors play no role in feature selection, which uses only the AtP$^*$ contribution to the answer-token log-odds. Alignment with them is therefore independent evidence that the selected features encode phonetic structure. The alignment is strongest for emotion, where the expected cue is RMS energy \citep{scherer2003vocal}. Top features reach $|r|=0.51$--$0.73$ on acted CREMA-D for Qwen and AF3. That same energy alignment carries onto VESUS at moderate strength (AF3 $0.51$, DeSTA up to $0.49$; Figure~\ref{fig:sae-mechanism-grid}). Intonation shows a different signature. On IViE, the top features of Qwen, Phi-4, and DeSTA correlate most with the expected F0-slope cue \citep{pierrehumbert1990meaning,ladd2008intonational}, but only at $|r|\approx0.3$. Yet their activations still separate question from statement clips into two clear clusters, most visibly for DeSTA (Figure~\ref{fig:sae-mechanism-grid}, left). Clean categorical separation despite weak F0-slope correlation implies these features respond to a more complex cue than F0 slope alone.

Causal sufficiency and acoustic interpretability are distinct claims, and they hold to different extents. Sufficiency is architecture-general: the clamp recovers the gated class on all four models (Table~\ref{tab:sfc-sufficiency}). Acoustic interpretability is narrower: strong correlates ($|r|>0.5$) appear only on emotion cells of Qwen and AF3 (Tables~\ref{tab:sfc-descriptors-qwen}--\ref{tab:sfc-descriptors-desta}). No Phi-4 feature exceeds $|r|=0.29$ on any descriptor (Table~\ref{tab:sfc-descriptors-phi4}), consistent with Phi-4's weaker audio-path prosodic signal (\S\ref{sec:study-1}).

Together, these results refine the dense recoverability result from \S\ref{sec:causal}. The underused prosodic signal can be recovered not only by a dense residual-stream edit, but also through a compact sparse subspace. On the emotion contrasts, that subspace contains features whose activations align with the expected energy-based acoustic cues while the intonation-side alignment is directionally consistent but weak.

%% file: tab_lens_ladder_verdicts.tex
\begin{table}[t]
\centering\scriptsize\setlength{\tabcolsep}{2.4pt}
\begin{tabular}{llcccl}
\toprule
Cell & Model & AUC ($L^*$) & $I_{\mathcal V}$ & \% rec. & verdict \\
\midrule
IViE Q/stmt & Qwen2.5-Omni & 0.82 (L26) & 0.32$\pm$0.24 & 36\% & F3 underuse \\
 & Phi-4-MM & 0.81 (L25) & 0.41$\pm$0.19 & 31\% & F3 underuse \\
 & AF3 & 0.83 (L24) & 0.23$\pm$0.28 & 30\% & F3 underuse \\
 & DeSTA2.5 & 1.00 (L18) & 1.00$\pm$0.00 & 0\% & F3 (complete) \\
\midrule
CREMA-D & Qwen2.5-Omni & 0.89 (L27) & 1.56$\pm$0.06 & 48\% & F3 underuse \\
 & Phi-4-MM & 0.61 (L29) & 0.93$\pm$0.09 & 23\% & F1 (\S\ref{sec:study-1}) \\
 & AF3 & 0.99 (L25) & 1.75$\pm$0.08 & 92\% & $\approx$reference \\
 & DeSTA2.5 & -- & -- & -- & masked (in-train) \\
\midrule
VESUS & Qwen2.5-Omni & 0.63 (L27) & 0.93$\pm$0.06 & 24\% & partial F2 \\
 & Phi-4-MM & 0.56 (L28) & 0.62$\pm$0.06 & 16\% & F1 (\S\ref{sec:study-1}) \\
 & AF3 & 0.84 (L25) & 1.55$\pm$0.04 & 43\% & F3 underuse \\
 & DeSTA2.5 & 0.81 (L32) & 1.49$\pm$0.05 & 59\% & F3 underuse \\
\bottomrule
\end{tabular}
\caption{Per-cell internal-representation and behavioral readouts across the four models. AUC = peak logit-lens AUC at layer $L^*$. $I_{\mathcal V}$ = predictive V-information at $L^*$ in bits $\pm$ bootstrap 95\% CI half-width (ceiling = label entropy: 1.0 bit for IViE, $\approx$2.0 bits for 4-class emotion; Appendix~\ref{app:vinfo}). Verdicts follow the \S\ref{sec:underuse} taxonomy and rest on the lens and ladder. The $1.00\pm0.00$ entry (DeSTA $\times$ IViE) reflects perfectly separable states, for which the bootstrap CI is degenerate.}
\label{tab:lens-ladder-verdicts}
\end{table}

%% file: tab_di_slopes_4model.tex
\begin{table}[t]
\centering\scriptsize\setlength{\tabcolsep}{3.5pt}
\begin{tabular}{lcccc}
\toprule
Contrast & Qwen & Phi-4 & AF3 & DeSTA \\
\midrule
IViE Q/stmt & $+1.34$ & $+1.84$ & $+2.26$ & $+0.29$ \\
CREMA-D h/s & $+1.76$ & $+0.64$ & $+3.20$ & \emph{masked} \\
CREMA-D n/a & $+2.01$ & $+0.48$ & $+2.57$ & \emph{masked} \\
VESUS h/s & $+0.95$ & $+0.47$ & $+2.04$ & $+1.83$ \\
VESUS n/a & $+1.09$ & $+0.31$ & $+2.80$ & $+2.92$ \\
\bottomrule
\end{tabular}
\caption{Direction-injection slopes at $L^*$ across the eighteen clean (model $\times$ contrast) cells. All slopes are positive with 95\% confidence intervals excluding zero (Table~\ref{tab:di-slope-ci}).}
\label{tab:di-slopes}
\vspace{-10px}
\end{table}

%% file: tab_patching_vesus.tex
\begin{table}[t]
\centering\scriptsize\setlength{\tabcolsep}{2pt}
\begin{tabular}{lcccc}
\toprule
VESUS & Qwen & Phi-4 & AF3 & DeSTA \\
\midrule
happy/sad     & $+0.9/{-}0.9$ & $+0.5/{-}0.6$ & $+3.3/{-}3.4$ & $+3.6/{-}3.8$ \\
neutral/angry & $+1.3/{-}1.3$ & $+0.5/{-}0.4$ & $+5.1/{-}5.2$ & $+3.6/{-}3.5$ \\
\bottomrule
\end{tabular}
\caption{Activation-patching $\Delta\log$-odds at $L^*$ on VESUS. All sixteen shifts take the predicted sign with bootstrap 95\% confidence intervals excluding zero, and effects are antisymmetric within each cell.}
\label{tab:patching}
\vspace{-10px}
\end{table}

%% file: sections/conclusion.tex
\section*{Conclusion}

Across four understanding-only audio-LLMs, the 
dominant explanation for prosody failures in our matched-content cells is \textbf{F3 underuse}. Prosodic information survives the audio path and is decodable in the late LLM stack, yet is only partially expressed in the model's answer. This signal is causally recoverable rather than merely probe-readable. Single-site interventions at the lens-peak layer shift the answer distribution in the predicted direction in every clean cell and drive the model toward the gated class in most, albeit as a directional bias rather than a selective restoration of the correct decision. A compact subspace of attribution-selected SAE features can drive the same recovery. The recurring bottleneck is therefore not hearing prosody but using it. Progress on expressive speech understanding may depend less on better audio encoders than on training and decoding choices that promote an already-available internal representation into behavior.

%% file: sections/limitations.tex
\section*{Limitations}

\paragraph{Models and scope.}

We probe four dense-transformer audio-LLMs (Qwen2.5-Omni-7B, Phi-4-MM, Audio-Flamingo-3, DeSTA2.5-Audio). MoE architectures (e.g.\ Qwen3-Omni \citep{qwen3omni}) break the single-residual-stream assumption our lens and SAE analyses rely on, and speech-to-speech models \citep{defossez2024moshi} additionally generate prosody. Whether F3 underuse transfers to either setting is open.

\paragraph{Cell coverage.}
F3 underuse holds in seven of the eleven clean (model $\times$ contrast) cells. The exceptions are systematic rather than random---Phi-4's matched-emotion cells inherit an upstream F1 weakness that the ladder cannot fully decompose, Qwen $\times$ VESUS pairs a weak internal code with weak behavior (partial F2), and AF3 $\times$ CREMA-D effectively reaches its text-cue reference---and training contamination forces masking of DeSTA $\times$ CREMA-D. We therefore treat F3 as the dominant recurring pattern, not a universal account.

\paragraph{Reference condition.}
The text+cue reference is a practical upper bound on the model's response to an explicit lexical cue, not a true ceiling on audio-driven behavior. A model that systematically downweights prosody relative to lexical content would produce the same baseline $<$ audio $<$ reference pattern.

\paragraph{Interventions.}
Direction injection shows that $L^*$ can move the output, but at the $\alpha$ values needed for full recovery, some negative-class clips also cross the decision boundary. In effect, the intervention biases the output toward a pole rather than selectively recovering the correct class. The scale of $\alpha$ is not a calibrated measure of natural gating strength (Appendix~\ref{app:causal-recovery}).

\paragraph{Logit lens.}
We use the basic logit lens rather than a tuned variant \citep{belrose2023eliciting}. Late-layer peaks are robust; intermediate-layer profiles should be read with the usual caveats. 

\paragraph{Acoustic correlates.}
Strong SAE acoustic alignment ($|r|>0.5$) is limited to the emotion cells of Qwen and AF3; DeSTA's VESUS correlations peak just below the threshold, the intonation cells are weak for every model, and Phi-4 shows no $|r|>0.5$ anywhere. Descriptors are hand-selected and uncorrected for multiple comparisons, so correlations are descriptive, not confirmatory.

%% file: appendix.tex
\section{Experimental Setup Details}
\label{app:setup}

\subsection{Audio Models}
\label{app:models}

Table~\ref{tab:app-models} lists all audio models used in the paper. The four audio-LLMs (Qwen2.5-Omni-7B, Phi-4-MM, Audio-Flamingo-3, and DeSTA2.5-Audio) are the probed systems. Six standalone speech encoders serve only as a calibration reference in \S\ref{sec:study-1}.

We choose Qwen2.5-Omni \citep{qwen25omni}, Phi-4-MM \citep{phi4mm}, Audio-Flamingo-3 \citep{audioflamingo3}, and DeSTA2.5-Audio \citep{desta25audio}. They are modern understanding-only audio-LLMs with diverse audio front ends (three Whisper-style towers and one conformer) and accessible hidden states. These properties allow the full probe ladder and causal/SAE interventions. Speech-to-speech models such as Moshi \citep{defossez2024moshi} are excluded since they additionally generate prosody, placing them outside our understanding-only scope. MoE systems such as Qwen3-Omni \citep{qwen3omni} are excluded because they break the single-residual-stream assumption that our lens and SAE analysis relies on. Finally, we treat SALMONN \citep{salmonn} and Qwen2-Audio \citep{qwen2audio} as related or secondary systems rather than main targets to keep the paper focused.

\begin{table*}[t]
\scriptsize
\centering
\setlength{\tabcolsep}{3pt}
\begin{tabular}{lllrl}
\toprule
Model & HuggingFace checkpoint & Architecture & \#~Params & Role \\
\midrule
\multicolumn{5}{l}{\emph{Probed audio-LLMs (\S\ref{sec:study-1}--\S\ref{sec:sfc})}} \\
\midrule
Qwen2.5-Omni-7B   & \texttt{Qwen/Qwen2.5-Omni-7B}                & Whisper-style audio tower + 28-layer LLM & 7\,B   & audio path \& LLM probed \\
Audio-Flamingo-3  & \texttt{nvidia/audio-flamingo-3-hf}          & Whisper-large-v3 + 28-layer Qwen2.5-7B LLM & 8.3\,B & audio path \& LLM probed \\
DeSTA2.5-Audio    & \texttt{DeSTA-ntu/DeSTA2.5-Audio-Llama-3.1-8B} & Whisper-large-v3 + 32-layer Llama-3.1-8B LLM & 8.7\,B & audio path \& LLM probed \\
Phi-4-MM          & \texttt{microsoft/Phi-4-multimodal-instruct} & Conformer audio tower + 32-layer LLM     & 5.6\,B & audio path \& LLM probed \\

\midrule
\multicolumn{5}{l}{\emph{Standalone speech encoders (\S\ref{sec:study-1} calibration only)}} \\
\midrule
Whisper-base.en   & \texttt{openai/whisper-base.en}              & encoder--decoder transformer (English)   & 74\,M  & encoder calibration \\
Whisper-medium    & \texttt{openai/whisper-medium}               & encoder--decoder transformer             & 769\,M & encoder calibration \\
Whisper-large-v2  & \texttt{openai/whisper-large-v2}             & encoder--decoder transformer             & 1.55\,B & encoder calibration \\
Whisper-large-v3  & \texttt{openai/whisper-large-v3}             & encoder--decoder transformer             & 1.55\,B & encoder calibration \\
WavLM-base        & \texttt{microsoft/wavlm-base}                & SSL transformer encoder                  & 95\,M  & encoder calibration \\
WavLM-large       & \texttt{microsoft/wavlm-large}               & SSL transformer encoder                  & 317\,M & encoder calibration \\
\bottomrule
\end{tabular}
\caption{All audio models used in the paper. Parameter counts are taken from the public model cards and are total architecture sizes, including text/decoder branches where applicable.}
\label{tab:app-models}
\end{table*}

\subsection{Dataset Routing across Studies}
\label{app:dataset-routing}

Table~\ref{tab:dataset-routing} shows which corpus is used in which study and for which diagnostic role. Audio-path probes in \S\ref{sec:study-1} use IViE for the Q/stmt contrast and the class-balanced 4-class emotion corpora (CREMA-D, ESD-English, and the held-out VESUS) for emotion. JL-Corpus appears only in the standalone-encoder calibration sweep (Appendix~\ref{app:probes}). For the lens-and-behavior analysis in \S\ref{sec:underuse}, the main claims are drawn from the three main cells (IViE Q/stmt, CREMA-D, and held-out VESUS, with in-training cells masked). ESD-English replicates the behavioral pattern under looser confound control in Appendix~\ref{app:underuse-supplement}; JL-Corpus is excluded there (in AF3's training data). Direction injection can be run on labeled contrasts, whereas activation patching requires matched-pair structure. In \S\ref{sec:causal}, both interventions are reported on all three corpora. Held-out VESUS serves as the main-text patching grid, and the CREMA-D and IViE cells appear in Appendix~\ref{app:causal-patching-cells}. The SAE analysis in \S\ref{sec:sfc} uses the same three corpora as mechanism cells, with \textbf{VESUS} additionally held out at the content level. For that corpus, the SAE is trained on a 70\% content-group split, and the sufficiency and acoustic-correlate results are reported on the disjoint held-out 30\%. Finally, the DeSTA $\times$ CREMA-D masking carries through \S\ref{sec:causal}--\S\ref{sec:sfc}.

\begin{table*}[t]
\scriptsize
\centering
\setlength{\tabcolsep}{4pt}
\begin{tabular}{lcccccc}
\toprule
 & \S\ref{sec:study-1} & \multicolumn{2}{c}{\S\ref{sec:underuse}} & \multicolumn{2}{c}{\S\ref{sec:causal}} & \S\ref{sec:sfc} \\
\cmidrule(lr){3-4}\cmidrule(lr){5-6}
Corpus & audio probe & lens & behavior & injection & patching & SAE \\
\midrule
IViE         & \checkmark & main & main & \checkmark & \checkmark\ & \checkmark \\
CREMA-D      & \checkmark & main & main & \checkmark & \checkmark\ & \checkmark \\
VESUS        & \checkmark & main & main & \checkmark & \checkmark & \checkmark\ \\
JL-Corpus    & App.~\ref{app:probes} & --- & --- & --- & --- & --- \\
ESD-English  & \checkmark & Appendix~\ref{app:underuse-supplement} & Appendix~\ref{app:underuse-supplement} & --- & --- & --- \\
\bottomrule
\end{tabular}
\caption{Per-study dataset assignment. ``main'' = appears in the main-text figures and tables; ``Appendix''\ = relegated to a supplementary appendix; ``---'' = not used as an analysis cell at that stage.}
\label{tab:dataset-routing}
\end{table*}
 
\subsection{Class Distributions per Corpus}
\label{app:dataset-distributions}

The clip counts in Table~\ref{tab:dataset-distributions} are filtered subsets of the published corpora, not the full releases. Per-corpus filtering is as follows:

\begin{itemize}
\setlength{\itemsep}{2pt}
\item \textbf{IViE.} We use a curated Q/stmt minimal-pair subset of the read-sentences task: 3 lexical sentence pairs read by 72 speakers across 6 UK dialects. Two are strict minimal pairs whose statement and question forms differ only in final intonation/punctuation (``He is on the lilo.''\,/\,``\ldots lilo?''; ``You remembered the lillies.''\,/\,``\ldots lillies?''). The third is a near-pair in which the subject pronoun also differs (``We live in Ealing.''\,/\,``You live in Ealing?''). Full coverage would be $72\times 3\times 2 = 432$ clips. Two readings are unavailable after filtering (one statement, one question), giving 430 clips (215 Q + 215 stmt) and 214 complete (Q, stmt) pairs. The other IViE speech styles (retold story, free conversation, map task) are not used.
\item \textbf{CREMA-D.} We use the 4-emotion subset $\{$angry, happy, sad, neutral$\}$ of the published 6-emotion corpus, dropping \emph{disgust} and \emph{fear}. Because the source corpus has fewer neutral clips than non-neutral clips, we use a neutral-limited balanced subset with 1{,}087 clips per emotion, for 4{,}348 clips total.
\item \textbf{VESUS.} We use the 4-emotion subset $\{$happy, sad, angry, neutral$\}$ of the 5-emotion corpus, dropping \emph{fearful}. VESUS reads a phonetically balanced, semantically neutral script of 250+ short phrases, each spoken by 10 actors in every emotion, giving 10{,}073 clips.
\item \textbf{JL-Corpus.} We use the 4-emotion subset of the primary emotion set, dropping the fifth primary emotion \emph{excited} as well as all 5 secondary emotions. That leaves 240 clips per emotion for 960 total.
\item \textbf{ESD-English.} We use the 4-emotion subset of the English half, dropping the fifth emotion \emph{surprise}. The Mandarin half of ESD is not used. This leaves 3{,}500 clips per emotion (10 speakers $\times$ 350 sentences).
\end{itemize}

All five filtered analysis subsets are balanced by construction. For CREMA-D, balance is achieved by limiting the non-neutral classes to the neutral-class count.

\begin{table}[t]
\small
\centering
\setlength{\tabcolsep}{4pt}
\begin{tabular}{lrrrrr}
\toprule
Corpus & Q/S  & sad & happy & angry & neutral \\
\midrule
IViE         & 215     & ---  & ---   & ---   & --- \\
CREMA-D      & ---     & 1{,}087 & 1{,}087 & 1{,}087 & 1{,}087 \\
VESUS        & ---     & 2{,}517 & 2{,}518 & 2{,}519 & 2{,}519 \\
JL-Corpus    & ---     & 240   & 240   & 240   & 240 \\
ESD-English  & ---     & 3{,}500 & 3{,}500 & 3{,}500 & 3{,}500 \\
\bottomrule
\end{tabular}
\caption{Per-class clip counts. IViE's two classes are question (Q) and statement (S). All corpora are balanced in the filtered subsets.}
\label{tab:dataset-distributions}
\end{table}

\subsection{Prompt Protocol Details}
\label{app:prompts}

\paragraph{System prompts.}
All four architectures are queried with their official chat templates, unmodified. Qwen2.5-Omni receives the canonical system message distributed with the model: ``You are Qwen, a virtual human developed by the Qwen Team, Alibaba Group, capable of perceiving auditory and visual inputs, as well as generating text and speech.'' DeSTA2.5-Audio likewise takes a short task-focusing system message, ``Focus on the audio clip and answer the question.'', following its recommended evaluation format. Phi-4-MM's speech-understanding template (\texttt{<|user|><|audio\_1|>\ldots<|end|><|assistant|>}) does not include a system turn by design. Audio-Flamingo-3 is queried with a single user turn carrying the audio and text prompt. For these two we do not add a system message.

\paragraph{Prompt counts per study.}
The IViE Q/stmt ladder in \S\ref{sec:underuse} uses $N{=}6$ balanced forced-choice prompts. These cover 3 phrasings $\times$ 2 pole orders (e.g.\ ``question or a statement'' vs.\ ``statement or a question''). Answers are aggregated by per-clip majority vote. Pole-order balancing controls for answer-order bias. The binary cells in \S\ref{sec:causal}--\S\ref{sec:sfc} (Q/stmt, happy/sad, and neutral/angry) use $N{=}5$ paraphrased forced-choice prompts each (Table~\ref{tab:prompts-binary}). In those later sections the 4-class emotion contrast is split into two 2-class cells. Meanwhile, the 4-class emotion evaluation in \S\ref{sec:underuse} uses $N{=}7$ prompts of mixed format (Table~\ref{tab:prompts-emotion4cls}). This mix is 3 free-form direct-answer prompts (\texttt{A1}--\texttt{A3}) plus 4 MCQ shuffles (\texttt{B1}--\texttt{B4}). A Latin-square design places each emotion at each letter position exactly once across the four runs, controlling for positional bias on the MCQ side.

\begin{table}[t]
\scriptsize
\centering
\setlength{\tabcolsep}{3pt}
\renewcommand{\arraystretch}{1.15}
\begin{tabular}{llp{0.65\columnwidth}}
\toprule
Contrast & ID & Prompt text \\
\midrule
Q/stmt & P0 & Is the utterance a question or a statement? Answer in one word: \\
       & P1 & Did the speaker ask a question or make a statement? Answer in one word: \\
       & P2 & Was that a question or a statement? Answer `question' or `statement': \\
       & P3 & Is the speaker asking something, or just making a statement? In one word, `question' or `statement': \\
       & P4 & Classify the speech act: `question' or `statement'. One word: \\
\midrule
Happy/sad & P0 & Does the speaker sound happy or sad? Answer in one word: \\
          & P1 & Is the speaker's tone happy or sad? Reply `happy' or `sad': \\
          & P2 & Classify the speaker's mood: `happy' or `sad'. One word: \\
          & P3 & Listening to this clip, would you say the speaker is happy or sad? In one word: \\
          & P4 & Is the emotion in the speaker's voice happy or sad? Answer in one word: \\
\midrule
Neutral/angry  & P0 & Does the speaker sound angry or neutral? Answer in one word: \\
         & P1 & Is the speaker's tone angry or neutral? Reply `angry' or `neutral': \\
         & P2 & Classify the speaker's mood: `angry' or `neutral'. One word: \\
         & P3 & Listening to this clip, would you say the speaker is angry or neutral? In one word: \\
         & P4 & Is the emotion in the speaker's voice angry or neutral? Answer in one word: \\
\bottomrule
\end{tabular}
\caption{Binary forced-choice prompt banks (5 paraphrases per contrast), used for all \S\ref{sec:underuse}--\S\ref{sec:sfc} binary cells.}
\label{tab:prompts-binary}
\end{table}

\begin{table}[t]
\scriptsize
\centering
\setlength{\tabcolsep}{3pt}
\renewcommand{\arraystretch}{1.15}
\begin{tabular}{lp{0.7\columnwidth}}
\toprule
ID & Prompt text \\
\midrule
\multicolumn{2}{l}{\emph{Direct prompts (free-form generation; substring-matched)}} \\
\midrule
A1 & What emotion is the speaker feeling? Answer in one word: \\
A2 & How does the speaker sound? Reply with one emotion: happy, sad, angry, or neutral: \\
A3 & In one word, what emotion is conveyed by the speaker's tone? \\
\midrule
\multicolumn{2}{l}{\emph{MCQ prompts (shared instruction template, Latin-square shuffles)}} \\
\midrule
\multicolumn{2}{p{0.92\columnwidth}}{Instruction template: ``Listen to the audio and identify the speaker's emotion. (A) \{A\} (B) \{B\} (C) \{C\} (D) \{D\}. Answer with just the letter A, B, C, or D:''} \\
\addlinespace
B1 & A = happy, B = sad, C = angry, D = neutral \\
B2 & A = sad, B = angry, C = neutral, D = happy \\
B3 & A = angry, B = neutral, C = happy, D = sad \\
B4 & A = neutral, B = happy, C = sad, D = angry \\
\bottomrule
\end{tabular}
\caption{4-class emotion prompt bank, used for \S\ref{sec:underuse}'s 4-class emotion evaluation. The Latin-square design guarantees each emotion appears at each letter position exactly once across B1--B4.}
\label{tab:prompts-emotion4cls}
\end{table}

\paragraph{Pole-token verification.}
Table~\ref{tab:pole-tokens} lists the per-contrast verbalizer sets used by the logit-lens readout at the answer position. Because the lens reads vocabulary logits rather than word probabilities, we apply a single-piece verification filter before using a candidate string in the main analysis. Candidate verbalizers, including leading-space variants, punctuation marks, and MCQ option letters, are tokenized in the answer-position form under all four model tokenizers. These tokenizers are Qwen2.5-Omni, Phi-4-MM, Audio-Flamingo-3 (Qwen2.5), and DeSTA2.5-Audio (Llama-3.1). Candidates that do not map to a single tokenizer piece in any of the four are excluded from the shared pole set. The retained candidates therefore define a first-token lens score consistently within each model. Class scores are computed as log-sum-exp over the retained pole-token logits, and the lens metric is the difference between the positive- and negative-pole scores. Cross-model comparisons are made at the level of AUC and intervention effect size, not by assuming identical logit calibration across vocabularies. This single-token filtering follows common practice in vocabulary-logit and steering workflows, where class or concept directions are defined over tokenizer-valid single-token verbalizers \citep{nanda2022transformerlens}.

\begin{table*}[t]
\small
\centering
\setlength{\tabcolsep}{3pt}
\begin{tabular}{lll}
\toprule
Contrast & Positive pole & Negative pole \\
\midrule
Q/stmt        & \texttt{\,question, \,Question, \,asking, \,query, ?} & \texttt{\,statement, \,Statement, \,stating, \,fact, .} \\
Happy/sad     & \texttt{\,happy, \,Happy, \,joyful} & \texttt{\,sad, \,Sad, \,upset} \\
Neutral/angry & \texttt{\,angry, \,Angry} & \texttt{\,neutral, \,Neutral} \\
\midrule
4-class direct (A1--A3) & \multicolumn{2}{l}{\texttt{\,happy, \,sad, \,angry, \,neutral} (one per class)} \\
4-class MCQ (B1--B4)    & \multicolumn{2}{l}{\texttt{\,A, \,B, \,C, \,D} (one per letter)} \\
\bottomrule
\end{tabular}
\caption{Verified pole-token sets used for the first-token logit-lens analysis.}
\label{tab:pole-tokens}
\end{table*}
 
\paragraph{Output parsing.}
Direct-answer prompts are decoded greedily, lowercased, and matched against class-specific synonym lists using longest-match and word-boundary rules. This prevents shorter class names from creating substring collisions, e.g., treating ``sadness'' as a valid sad synonym rather than as an accidental match to ``sad''. MCQ prompts are parsed by locating the first standalone option marker, such as \texttt{A}, \texttt{(A)}, or \texttt{A.}, and the corresponding patterns for B/C/D. Markers are bounded by non-letter characters so that letters inside words such as ``Answer'' are not matched. The selected option is then mapped to the emotion label assigned under the active Latin-square shuffle. Generations with no valid class match are counted as incorrect in all main analyses. Finally, the lens readout uses the log-sum-exp pole scores defined in Table~\ref{tab:pole-tokens}.

\section{Audio-Path Probe Details}
\label{app:probes}

\paragraph{Probe architecture.}
The audio-path probes in \S\ref{sec:study-1} use a \textbf{single-layer linear head} on top of frozen encoder representations, following the standard probing-classifier protocol \citep{belinkov2022probing,yang2021superb}:
\[
\hat y = \mathrm{softmax}(Wh+b),
\]
where $W \in \mathbb{R}^{n_{\text{classes}}\times d_{\text{enc}}}$. For each (encoder, layer) pair, we extract per-frame hidden states for each clip and mean-pool over valid acoustic frames only. Padded frames, when present, are excluded using the model-specific attention mask. The resulting pooled vector $h$ is used as input to the linear probe. For audio-LLMs, the PROJ point denotes the post-projector representation passed to the LLM token stack, after the model-specific pooling/projection operations.

The linear head is trained on the pooled vectors with Adam (lr $=10^{-3}$, weight decay $=10^{-3}$) and cross-entropy loss. Training uses early stopping on a 10\% in-fold validation split (patience 30 epochs, maximum 300 epochs). Probe weights are re-initialized independently for every fold and layer. Cross-validation is speaker-disjoint leave-one-group-out, with the group definition adapted per corpus. IViE (72 speakers $\to$ 72 folds; IViE has many small per-speaker sets) and VESUS (10 actors $\to$ 10 folds) use per-speaker leave-one-speaker-out. So do JL-Corpus (4 actors $\to$ 4 folds) and ESD-English (10 speakers $\to$ 10 folds). CREMA-D uses 10-fold leave-one-bucket-out, where the 91 actors are deterministically bucketed by \texttt{actor\_id\,\%\,10}. This reduces the full encoder $\times$ layer sweep from 91 folds to 10 folds while preserving the key constraint: no speaker appears in both train and test of any fold.

We report weighted accuracy (WA), computed over the held-out clips in each fold; equivalently, WA is support-weighted per-class recall,
\[
\mathrm{WA}
=
\sum_c \frac{n_c}{N}
\frac{\mathrm{TP}_c}{n_c},
\]
which reduces to ordinary accuracy on balanced binary cells. Here \(n_c\) is the number of held-out clips with true class \(c\), and \(N\) is the total number of test clips across all classes. The count \(\mathrm{TP}_c\) is the number of those clips correctly predicted as class \(c\). Fold-level scores are averaged across folds. Each trained-probe result is compared against a random-initialized baseline. The baseline uses the same input pipeline, pooling, probe architecture, optimization settings, and cross-validation splits, but replaces the encoder/audio-tower weights with a fixed-seed random initialization. It therefore measures how much of the probe score can be obtained from architecture, preprocessing, and dataset statistics alone, rather than from learned speech representations.

\begin{figure*}[t]
\centering
\includegraphics[width=\textwidth]{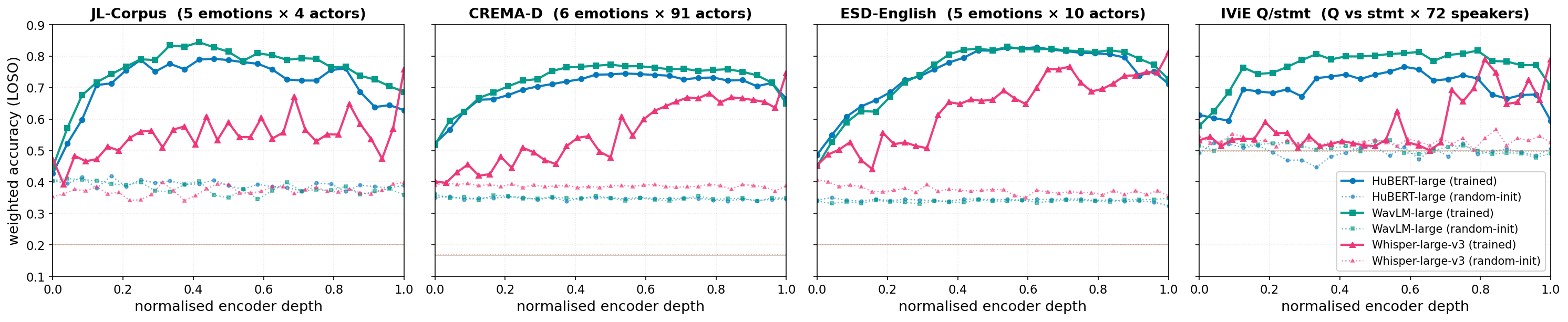}
\caption{Calibration of the probing protocol on standalone speech encoders. Layer-wise weighted accuracy (speaker-disjoint LOSO) of the single-layer linear probe on a representative self-supervised encoder across the three emotion corpora and IViE Q/stmt. Solid = trained encoder, dotted = random-initialized baseline.}
\label{fig:probes-ssl}
\end{figure*}

\paragraph{Calibration against published speech-emotion probing.}
The encoder probe sweeps in this appendix recover the qualitative pattern reported in recent speech-emotion probing benchmarks. Stronger pretrained encoders yield better linear decodability on categorical emotion corpora, and prosodic categories remain recoverable from frozen self-supervised speech representations \citep{ma2024emobox}. Figure~\ref{fig:probes-ssl} applies our single-layer probe to a representative encoder from HuBERT-large, WavLM-large, and the supervised Whisper-large-v3 tower on the three emotion corpora and IViE Q/stmt. On the emotion corpora, decodability rises through the encoder and every model separates clearly from its random-initialized baseline, reproducing the expected speech-emotion-recognition ordering. This serves as a calibration for the \S\ref{sec:study-1} audio-LLM probes. The protocol thus behaves as expected on these standalone encoders. We therefore apply the identical protocol to the audio-LLM towers and projectors (\S\ref{sec:study-1}, Figure~\ref{fig:audio-path-probes}).

\input{tab_audio_path_numbers}

\paragraph{Projector readout per model.}
The PROJ column is the representation each model hands to its LLM token stack. For Qwen2.5-Omni and Audio-Flamingo-3, this is the post-projector output of the Whisper-style audio tower. DeSTA2.5-Audio hands over the Q-Former \texttt{prompt\_output}, and Phi-4-MM uses the audio projector on the conformer encoder. All pooling is mask-aware (Appendix~\ref{app:probes}, ``Probe architecture''); padded frames are excluded.

\paragraph{Random-init baseline is a within-panel reference.}
The random-initialized baseline is generated per extractor and is not identical across models. Specifically, the AF3 and DeSTA towers use a seeded \texttt{xavier\_uniform} re-initialization, while the Qwen and Phi-4 extractors use \texttt{kaiming\_normal}. The resulting random-feature quality, and hence the absolute dotted baseline, therefore differs across panels. We accordingly read F1 from the trained-minus-random \emph{lift} within each panel rather than comparing absolute random-init WA across models. On the balanced 4-class emotion cells the random-init projector baseline is itself well above the 0.25 class prior (typically 0.38--0.60). This is because architecture, pooling, and dataset statistics alone carry some signal. The trained lift over this baseline is the F1 quantity of interest.

\paragraph{Training-data overlap.}
Two of the emotion corpora overlap with model training data and are \emph{masked} in Table~\ref{tab:audio-path-numbers}. CREMA-D and ESD-English are in DeSTA2.5's instruction-tuning mix, so we mask DeSTA's ESD and CREMA-D cells and exclude them from cross-model comparison. DeSTA's frozen Whisper-large-v3 tower is not itself trained on these corpora, so its layer-wise encoding is uncontaminated; we nonetheless mask the full cells, including the projector, out of caution. VESUS is held out for all four models and is the clean cross-model emotion comparison; IViE Q/stmt is synthetic matched-content. AF3's training overlap (TESS, JL-Corpus) does not include any \S\ref{sec:study-1} corpus, so AF3 is clean on IViE, CREMA-D, ESD, and VESUS.
 
\section{Text-Input Control for the Three-Condition Ladder}
\label{app:text-control}

\paragraph{Template.}
Both text conditions use the same wrapper, followed by the same per-contrast prompt from Appendix~\ref{app:prompts}:
\textbf{no-cue:} \texttt{The speaker says: ``\{transcript\}''};
\textbf{with-cue:} \texttt{The speaker says \{cue\}: ``\{transcript\}''}.
The cue is a short parenthetical phrase naming the prosodic state, lexicalized per contrast (Table~\ref{tab:text-cue-lex}).

\begin{table}[h]
\scriptsize
\centering
\begin{tabular}{lll}
\toprule
Contrast & Positive cue & Negative cue \\
\midrule
Q/stmt   & \texttt{(asking a question)}   & \texttt{(making a statement)} \\
\midrule
\multicolumn{3}{l}{\emph{4-class emotion (one cue per class)}} \\
\multicolumn{3}{l}{\texttt{(with happiness)}, \texttt{(with sadness)},
\texttt{(with anger)}, \texttt{(neutrally)}} \\
\bottomrule
\end{tabular}
\caption{State-naming cues used to construct the text+cue reference. Locked per pre-registration.}
\label{tab:text-cue-lex}
\end{table}

\paragraph{Transcript construction.}
Transcripts are drawn from the per-corpus index files and rendered so that terminal punctuation does not reveal the target label. For the emotion corpora (CREMA-D, VESUS, ESD-English), the transcripts are used as given, since all sentences end with the same punctuation regardless of emotion. In the IViE Q/stmt case, punctuation would reveal the label (`?' vs.\ `.'), so each item is mapped to a hand-written punctuation-free canonical sentence string.

\paragraph{Why some references are below one.}
\label{app:ivie-ceiling-note}

In Figure~\ref{fig:results-ladder}, several emotion ceilings sit slightly below 1.0 (e.g.\ Qwen $\times$ VESUS 0.98, DeSTA $\times$ VESUS 0.91). Even when the emotion is supplied verbally as a parenthetical cue, the LLM does not express it correctly on every clip. The \% recovery denominator is therefore the baseline-to-reference range actually available to the model under a text cue. A sub-1.0 reference rescales the available range but does not bias the F3 diagnosis, since the ordering baseline $<$ audio $<$ reference still holds.

\section{Lens and Ladder Details}
\label{app:underuse-details}

\subsection{Layer-wise Lens Curves for the Main Cells}
\label{app:lens-curves}

\begin{figure}[t]
\centering
\includegraphics[width=\columnwidth]{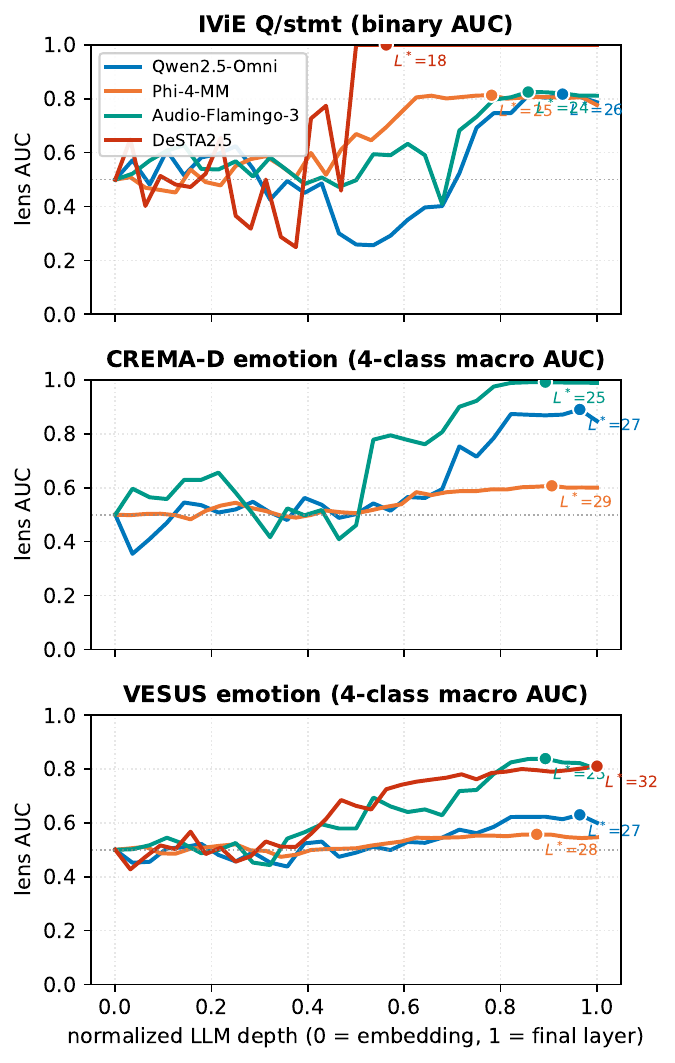}
\caption{Layer-wise lens AUC, four architectures overlaid: \textbf{\textcolor[HTML]{0077BB}{Qwen2.5-Omni}}, \textbf{\textcolor[HTML]{EE7733}{Phi-4-MM}}, \textbf{\textcolor[HTML]{009988}{Audio-Flamingo-3}}, \textbf{\textcolor[HTML]{CC3311}{DeSTA2.5}}. Top: IViE Q/stmt (binary AUC on the prosodic-pole tokens). Middle: CREMA-D emotion (4-class macro AUC; DeSTA masked, in-training). Bottom: held-out VESUS (4-class macro AUC). Dots mark $L^*$.}
\label{fig:results-lens-auc}
\end{figure}

Figure~\ref{fig:results-lens-auc} shows the per-layer logit-lens AUC behind the $L^*$ peaks reported in Table~\ref{tab:lens-ladder-verdicts}. Lens AUC rises late in the stack on every strongly decodable cell, consistent with the answer-position prosodic code being assembled in the upper third of the LLM.

\subsection{V-information Estimation Details}
\label{app:vinfo}

\paragraph{Definition.}
Predictive V-information \citep{xu2020vinformation} measures the label information extractable from a representation by a \emph{bounded} predictor family $\mathcal V$. With $X = h_{L^*}$ the answer-position residual stream at the lens-peak layer and $Y$ the prosodic label,
\[
I_{\mathcal V}(X \to Y) \;=\; H_{\mathcal V}(Y \mid \emptyset) \;-\; H_{\mathcal V}(Y \mid X),
\]
where $H_{\mathcal V}(Y \mid X) = \inf_{f \in \mathcal V} \mathbb E\!\left[-\log_2 f[x](y)\right]$ is the best held-out log-loss achievable in the family. Similarly, $H_{\mathcal V}(Y \mid \emptyset)$ is the same under the label prior alone. Setting $\mathcal V$ to all functions recovers Shannon mutual information. Restricting it makes the measure sensitive to \emph{how} the information is encoded. A bijective scrambling of $X$ leaves $I(X;Y)$ unchanged but destroys $I_{\mathcal V}$, which is the property we want for a usable-representation claim. Empirically, $\hat I_{\mathcal V}$ is the mean per-clip log-likelihood ratio of the probe over the prior, in bits, on held-out clips.

\paragraph{Instantiation.}
We take $\mathcal V$ to be L2-regularized linear softmax classifiers on standardized features, matching the linearity of the model's own unembedding. Information measured this way lies in the model's access basis, sharpening F3 (represented and linearly accessible, yet unused) against F2 (buried non-linearly). The regularization constant is selected by validation log-loss ($C=0.01$). Evaluation uses a 20\% held-out split and a 2{,}000-rep bootstrap for 95\% confidence intervals. Here the ceiling is the label entropy: $\approx$2.0 bits for the balanced 4-class emotion cells, 1.0 bit for IViE Q/stmt. Two controls accompany every cell. A shuffled-label control task \citep{hewitt2019control} re-estimates $I_{\mathcal V}$ with permuted labels and yields $\approx$0 bits, so selectivity $\approx$ the full reported value. The bits reflect structure, not probe capacity. Additionally, a speaker-disjoint split (train/test by speaker) checks speaker leakage.
 
\subsection{Supplementary Emotion Corpora: ESD-English}
\label{app:underuse-supplement}

The main text reports IViE Q/stmt, CREMA-D, and VESUS. Here we report the same lens-and-ladder protocol on one additional matched-content emotion corpus, ESD-English (DeSTA $\times$ ESD is masked: in-training).

The behavioral pattern generalizes (Table~\ref{tab:supp-emotion-ladder}). In every clean cell, the audio condition improves over the text-no-cue floor but remains below the text+cue ceiling, and the model gradient from the main text reappears. AF3 recovers the most (60\%), while Qwen and Phi-4 recover 17--26\%. The lens evidence is weaker than on the main cells, however. Only AF3 $\times$ ESD clears the AUC $\geq 0.75$ confirmation threshold, so the Qwen and Phi-4 cells stay partial F2 (lens $\leq 0.65$) rather than confirmed F3 cases.

This scoping matters because the corpora differ in diagnostic strength. CREMA-D and VESUS provide the most diagnostic matched-emotion settings in the main text, with large speaker pools and repeated transcripts across emotion conditions. ESD-English also contains parallel emotion recordings, but its corpus-specific recording properties may dilute the answer-position emotion signal measured by the lens. We therefore use it to test whether the floor--audio--ceiling behavioral pattern generalizes, not as primary evidence for a fully localized F3 mechanism.

\begin{table*}[t]
\centering
\small
\begin{tabular}{llcccccc}
\toprule
Corpus & Model & lens AUC ($L^*$) & Floor & Audio & Ceiling & \% ladder & Verdict \\
\midrule
ESD-English & Qwen2.5-Omni & 0.63 & 0.159 & 0.371 & 0.970 & 26\% & partial F2 \\
            & Phi-4-MM     & 0.53 & 0.128 & 0.266 & 0.963 & 17\% & partial F2 \\
            & AF3          & 0.90 & 0.207 & 0.680 & 0.991 & 60\% & F3 underuse \\
            & DeSTA2.5     & \multicolumn{6}{c}{\emph{in-training data (masked)}} \\
\bottomrule
\end{tabular}
\caption{Supplementary emotion cells on ESD-English under the same protocol as the main text. In-training (model $\times$ corpus) cells are masked. Verdicts follow the \S\ref{sec:underuse} taxonomy.}
\label{tab:supp-emotion-ladder}
\end{table*}

\section{Causal Intervention Details}
\label{app:causal}

This appendix provides implementation details for the causal interventions in \S\ref{sec:causal}. The details cover direction-injection per-$\alpha$ sweeps with slope confidence intervals (\S\ref{app:causal-injection}) and the CREMA-D and IViE activation-patching cells (\S\ref{app:causal-patching-cells}). Behavior-level recovery sweeps for all eighteen clean cells appear in \S\ref{app:causal-sweep}. They also cover the interpretation of the $\alpha$ scale (\S\ref{app:causal-recovery}) and activation-patching layer/position ablations (\S\ref{app:causal-patching}).

\subsection{Direction-Injection: Per-$\alpha$ Sweeps and Slope Confidence Intervals}
\label{app:causal-injection}

Table~\ref{tab:di-slope-ci} reports the direction-injection slope with its 95\% confidence interval for every main-text cell. All eighteen intervals exclude zero, so the positive-sign result in Table~\ref{tab:di-slopes} is robust on every (architecture $\times$ contrast) combination.
 
\input{tab_di_slope_ci_4model}

\subsection{Activation Patching: CREMA-D and IViE Cells}
\label{app:causal-patching-cells}

Table~\ref{tab:patching-cremad-ivie} reports the donor activation-patching cells on CREMA-D and IViE for all four models, complementing the main-text VESUS grid (Table~\ref{tab:patching}). All twenty entries take the predicted sign with bootstrap 95\% confidence intervals excluding zero.

\begin{table}[t]
\centering
\scriptsize
\begin{tabular}{llcrr}
\toprule
Model & Cell & $n$ & $-\!\!\to\!+$ & $+\!\!\to\!-$ \\
\midrule
Qwen2.5-Omni & IViE Q/stmt              & 214 & $+1.15$ & $-1.15$ \\
Qwen2.5-Omni & CREMA-D happy/sad        & 1{,}087 & $+3.23$ & $-3.17$ \\
Qwen2.5-Omni & CREMA-D neutral/angry    & 1{,}087 & $+3.37$ & $-3.38$ \\
Phi-4-MM     & IViE Q/stmt              & 214 & $+1.69$ & $-1.80$ \\
Phi-4-MM     & CREMA-D happy/sad        & 1{,}087 & $+1.00$ & $-1.08$ \\
Phi-4-MM     & CREMA-D neutral/angry    & 1{,}087 & $+0.54$ & $-0.47$ \\
Audio-Flamingo-3 & IViE Q/stmt          & 150 & $+1.59$ & $-1.64$ \\
Audio-Flamingo-3 & CREMA-D happy/sad    & 150 & $+8.17$ & $-8.75$ \\
Audio-Flamingo-3 & CREMA-D neutral/angry & 150 & $+7.95$ & $-8.34$ \\
DeSTA2.5     & IViE Q/stmt              & 150 & $+7.15$ & $-7.60$ \\
DeSTA2.5     & CREMA-D h/s, n/a         & \multicolumn{3}{c}{\emph{masked (in-training)}} \\
\bottomrule
\end{tabular}
\caption{Activation-patching $\Delta\log$-odds at $L^*$ on CREMA-D and IViE (canonical prompt). All twenty entries take the predicted sign, with bootstrap 95\% confidence intervals excluding zero.}
\label{tab:patching-cremad-ivie}
\end{table}

\subsection{Behavioral Recovery Sweeps}
\label{app:causal-sweep}

Table~\ref{tab:di-recall-sweep} reports the behavior-level $\alpha$-sweep for all eighteen clean cells under the five-prompt majority vote. At each $\alpha$, a clip's decoded answer is the majority, across the five paraphrase prompts, of the sign of the prosodic answer-token log-odds. Positive-class recall saturates within the tested grid (through $\alpha{=}8$) in fifteen of the eighteen cells, and the three exceptions are systematic. Phi-4 $\times$ VESUS happy/sad starts near ceiling (0.972), leaving little to recover. Meanwhile, Phi-4 $\times$ VESUS neutral/angry rises only to 0.27 at $\alpha{=}5$ (0.65 at $\alpha{=}8$), consistent with Phi-4's weak VESUS code (\S\ref{sec:study-1}). DeSTA $\times$ IViE never crosses the decision boundary within the tested grid. Its slope ($+0.29$, Table~\ref{tab:di-slope-ci}) is too small to overcome the model's statement-pole offset at these $\alpha$ values. This is consistent with its extreme behavioral bias (\S\ref{sec:underuse}).

\begin{table*}[h]
\small
\centering
\setlength{\tabcolsep}{3pt}
\begin{tabular}{llrrrrrr}
\toprule
Model & Cell & $\alpha{=}0$ & $\alpha{=}1$ & $\alpha{=}2$ & $\alpha{=}3$ & $\alpha{=}5$ & $\alpha{=}8$ \\
\midrule
Qwen2.5-Omni & IViE Q/stmt           & 0.138 & 0.738 & 0.954 & \textbf{1.000} & 1.000 & 1.000 \\
Qwen2.5-Omni & CREMA-D happy/sad     & 0.725 & \textbf{1.000} & 1.000 & 1.000 & 1.000 & 1.000 \\
Qwen2.5-Omni & CREMA-D neutral/angry & 0.366 & 0.759 & \textbf{1.000} & 1.000 & 1.000 & 1.000 \\
Qwen2.5-Omni & VESUS happy/sad       & 0.388 & 0.828 & 0.894 & 0.930 & 0.995 & \textbf{1.000} \\
Qwen2.5-Omni & VESUS neutral/angry   & 0.168 & 0.238 & 0.337 & 0.512 & \textbf{1.000} & 1.000 \\
\midrule
Phi-4-MM     & IViE Q/stmt           & 0.015 & 0.092 & 0.369 & 0.800 & \textbf{1.000} & 1.000 \\
Phi-4-MM     & CREMA-D happy/sad     & 0.992 & \textbf{1.000} & 1.000 & 1.000 & 1.000 & 1.000 \\
Phi-4-MM     & CREMA-D neutral/angry & 0.448 & 0.709 & 0.916 & 0.990 & \textbf{1.000} & 1.000 \\
Phi-4-MM     & VESUS happy/sad       & 0.972 & 0.974 & 0.974 & 0.979 & 0.985 & 0.992 \\
Phi-4-MM     & VESUS neutral/angry   & 0.118 & 0.134 & 0.155 & 0.185 & 0.272 & 0.647 \\
\midrule
Audio-Flamingo-3 & IViE Q/stmt           & 0.354 & 0.692 & \textbf{1.000} & 1.000 & 1.000 & 1.000 \\
Audio-Flamingo-3 & CREMA-D happy/sad     & 0.945 & \textbf{1.000} & 1.000 & 1.000 & 1.000 & 1.000 \\
Audio-Flamingo-3 & CREMA-D neutral/angry & 0.958 & \textbf{1.000} & 1.000 & 1.000 & 1.000 & 1.000 \\
Audio-Flamingo-3 & VESUS happy/sad       & 0.530 & 0.959 & \textbf{1.000} & 1.000 & 1.000 & 1.000 \\
Audio-Flamingo-3 & VESUS neutral/angry   & 0.614 & 0.896 & \textbf{1.000} & 1.000 & 1.000 & 1.000 \\
\midrule
DeSTA2.5     & IViE Q/stmt           & 0.000 & 0.000 & 0.000 & 0.000 & 0.000 & 0.000 \\
DeSTA2.5     & CREMA-D h/s, n/a      & \multicolumn{6}{c}{\emph{masked (in-training)}} \\
DeSTA2.5     & VESUS happy/sad       & 0.550 & 0.796 & 0.968 & \textbf{1.000} & 1.000 & 1.000 \\
DeSTA2.5     & VESUS neutral/angry   & 0.597 & 0.744 & 0.979 & \textbf{1.000} & 1.000 & 1.000 \\
\bottomrule
\end{tabular}
\caption{Positive-class recall under raw residual-stream direction injection at $L^*$ across the eighteen clean cells, evaluated on a 30\% held-out TEST split per cell under the five-prompt majority vote (\S\ref{sec:setup-prompts}): at each $\alpha$ a clip's decoded class is the majority across the five paraphrase prompts of the sign of the answer-token log-odds. The class-mean direction $d{=}\mu^+-\mu^-$ is fit on the remaining 70\% TRAIN split. Bold marks the per-cell working $\alpha$ (smallest $\alpha$ at which recall$_+$ reaches $1.000$); the three cells without bold do not saturate in the tested range (see text). All cells vote over the full five-prompt bank except AF3$\times$IViE, which votes over four (the canonical prompt's per-clip log-odds were not retained for that cell).}
\label{tab:di-recall-sweep}
\end{table*}

\subsection{Interpreting and Scaling the $\alpha$ Intervention}
\label{app:causal-recovery} 

The intervention strength $\alpha$ is an editing parameter, not a direct measure of how strongly the model gates prosody. Consider the edit $h_{\mathrm{last}} \leftarrow h_{\mathrm{last}} + \alpha d$. Here the scale of $\alpha$ depends on the norm of the class-mean direction $d$ and the variability of states within each class. It also depends on the prompt and the distance from the current state to the answer-token decision boundary. Thus, a larger working $\alpha$ should not be read as evidence for stronger natural gating. As a geometric reference, Table~\ref{tab:di-norm-ratio} reports $\|d\|$, the median $\|h_{\mathrm{last}}\|$, and their ratio for the original Qwen and Phi-4 cells. At $\alpha=1$, the added displacement on these cells is $0.076$--$0.187$ of the median state norm. This shows that recovery is produced by a finite and measurable late-state edit, rather than providing a calibrated estimate of the model's natural gating strength.

\begin{table*}[h]
\small
\centering
\setlength{\tabcolsep}{4pt}
\begin{tabular}{llrrr}
\toprule
Model & Cell & $\|d\|$ & $\|h_{\mathrm{last}}\|_{\mathrm{med}}$ & $\|d\|/\|h_{\mathrm{last}}\|$ \\
\midrule
Qwen2.5-Omni & IViE Q/stmt            & $31.9$ & $329$ & $0.097$ \\
Qwen2.5-Omni & CREMA-D happy/sad      & $61.5$ & $330$ & $0.187$ \\
Qwen2.5-Omni & CREMA-D neutral/angry  & $61.8$ & $331$ & $0.187$ \\
Phi-4-MM     & IViE Q/stmt            & $15.5$ & $155$ & $0.100$ \\
Phi-4-MM     & CREMA-D happy/sad      & $11.5$ & $151$ & $0.076$ \\
Phi-4-MM     & CREMA-D neutral/angry  & $12.6$ & $150$ & $0.084$ \\
\bottomrule
\end{tabular}
\caption{Geometric scale of the direction-injection intervention (original Qwen/Phi-4 cells). $\|d\|$ is the L2 norm of the class-mean direction $d=\mu^+-\mu^-$. Likewise, $\|h_{\mathrm{last}}\|_{\mathrm{med}}$ is the median L2 norm of the answer-position residual stream at $L^*$ over the union of positive- and negative-class clips. The ratio gives the fractional displacement applied at $\alpha{=}1$.}
\label{tab:di-norm-ratio}
\end{table*}

\subsection{Activation Patching: Layer and Position Ablations}
\label{app:causal-patching}

Table~\ref{tab:causal-patching} reports activation-patching effects under three settings. The three settings are an early-layer null control ($L{=}1$, answer position), the main targeted intervention ($L^*$, answer position), and a full-sequence control ($L^*$, all positions). For Qwen, $L^*=26$, while for Phi-4 we use a shared $L^*=25$ across cells to keep the patching and clamp pipelines comparable across contrasts. This is the IViE Q/stmt lens peak, while CREMA-D emotion peaks earlier at $L=22$ (Figure~\ref{fig:results-lens-auc}). The early-layer control gives near-zero effects, showing that the result is not a generic consequence of swapping hidden states. In contrast, the full-sequence control can move the output, but it does not consistently preserve the clean bidirectional antisymmetry of the answer-position patch. This is expected when many token states are replaced. It can produce the opposite shift, further localizing the causal locus to the answer position at $L^*$.

\begin{table*}[t]
\scriptsize
\centering
\setlength{\tabcolsep}{3pt}
\begin{tabular}{llccc}
\toprule
 & & \multicolumn{1}{c}{$L{=}1$, last} & \multicolumn{1}{c}{$L^*$, last} & \multicolumn{1}{c}{$L^*$, full} \\
\cmidrule(lr){3-3}\cmidrule(lr){4-4}\cmidrule(lr){5-5}
Model & Cell & $\Delta_{n\to p}$ / $\Delta_{p\to n}$ & $\Delta_{n\to p}$ / $\Delta_{p\to n}$ & $\Delta_{n\to p}$ / $\Delta_{p\to n}$ \\
\midrule
Qwen2.5-Omni & IViE Q/stmt
& $-0.00_{[-0.01,0.00]}$ / $+0.00_{[-0.00,0.01]}$
& $+1.15_{[0.96,1.35]}$ / $-1.15_{[-1.34,-0.96]}$
& $+1.15_{[0.79,1.53]}$ / $-0.18_{[-0.50,0.13]}$ \\

Qwen2.5-Omni & CREMA-D happy/sad
& $-0.00_{[-0.00,0.00]}$ / $+0.00_{[-0.00,0.01]}$
& $+3.23_{[3.15,3.32]}$ / $-3.17_{[-3.25,-3.08]}$
& $+1.12_{[0.73,1.50]}$ / $-1.52_{[-2.02,-1.05]}$ \\

Qwen2.5-Omni & CREMA-D neutral/angry
& $+0.00_{[-0.00,0.01]}$ / $+0.00_{[-0.00,0.01]}$
& $+3.37_{[3.27,3.48]}$ / $-3.38_{[-3.49,-3.28]}$
& $+3.23_{[2.72,3.78]}$ / $-1.20_{[-1.57,-0.84]}$ \\

Phi-4-MM & IViE Q/stmt
& $+0.00_{[-0.02,0.02]}$ / $+0.00_{[-0.01,0.02]}$
& $+1.69_{[1.40,2.00]}$ / $-1.80_{[-2.12,-1.49]}$
& $+3.19_{[2.60,3.79]}$ / $+1.65_{[1.07,2.21]}$ \\

Phi-4-MM & CREMA-D happy/sad
& $-0.00_{[-0.02,0.02]}$ / $-0.00_{[-0.02,0.01]}$
& $+1.00_{[0.90,1.09]}$ / $-1.08_{[-1.18,-0.98]}$
& $-0.10_{[-0.36,0.14]}$ / $-2.89_{[-3.34,-2.46]}$ \\

Phi-4-MM & CREMA-D neutral/angry
& $+0.02_{[-0.00,0.04]}$ / $-0.01_{[-0.02,0.01]}$
& $+0.54_{[0.47,0.62]}$ / $-0.47_{[-0.54,-0.39]}$
& $-0.37_{[-0.85,0.12]}$ / $-0.46_{[-0.74,-0.18]}$ \\
\bottomrule
\end{tabular}
\caption{Activation-patching $\Delta\log$-odds under layer and position controls. The $L^*$ answer-position column is the main intervention; subscripts give 95\% bootstrap confidence intervals.}
\label{tab:causal-patching}
\end{table*}

\section{SAE Training, AtP$^*$, and Clamp Details}
\label{app:sfc}

This appendix provides the implementation details and supporting
analyses for the sparse-feature intervention in \S\ref{sec:sfc}:
SAE training, AtP$^*$ feature attribution, $S_{0.95}$ selection,
the clamp intervention, and the acoustic-descriptor analyses.

\subsection{SAE Representation and Clamp Intervention}
\label{app:sfc-clamp}
For each clip, we take the answer-position residual stream $h_{\mathrm{last}}\in\mathbb{R}^d$ at the lens-peak layer $L^*$. For Qwen2.5-Omni, $d=3584$ and $L^*=26$; for Phi-4-MM, $d=3072$ and $L^*=25$; for Audio-Flamingo-3, $d=3584$ and $L^*=26$; for DeSTA2.5, $d=4096$ and $L^*=29$. A TopK SAE encodes this state as
\[
f_{\mathrm{pre}} = W_{\mathrm{enc}}h_{\mathrm{last}}+b_{\mathrm{enc}},
f=\mathrm{TopK}_{k}(f_{\mathrm{pre}}),
\]
with $k=64$. With expansion factor $8$, the feature dimension is $D=8d$ per model: 28{,}672 for Qwen and AF3, 24{,}576 for Phi-4, and 32{,}768 for DeSTA.

Let $S\equiv S_{0.95}$ be the attribution-selected feature set defined below. For each cell, positive- and negative-class feature-activation means, $\mu^{+}_{\mathrm{act}}$ and $\mu^{-}_{\mathrm{act}}$, are computed on the train split. The sparse clamp edits only features in $S$:
\[
\tilde f_{j} =
\begin{cases}
f_j+\alpha(\mu^{+}_{\mathrm{act},j}-\mu^{-}_{\mathrm{act},j}) & j\in S,\\
f_j & j\notin S.
\end{cases}
\]
TopK is not re-applied after the clamp, so a previously zero coordinate in $S$ may become active. This edited feature vector is then decoded back into residual-stream space,
\[
\hat h = W_{\mathrm{dec}}\tilde f+b_{\mathrm{dec}},
\]
and $\hat h$ overwrites the answer-position residual stream at $L^*$. From there, the forward pass continues normally through the remaining layers. No other token position, layer, or non-$S$ feature coordinate is modified.


\subsection{TopK SAE Training}
\label{app:sfc-training}
\begin{figure}[t]
\centering
\includegraphics[width=\columnwidth]{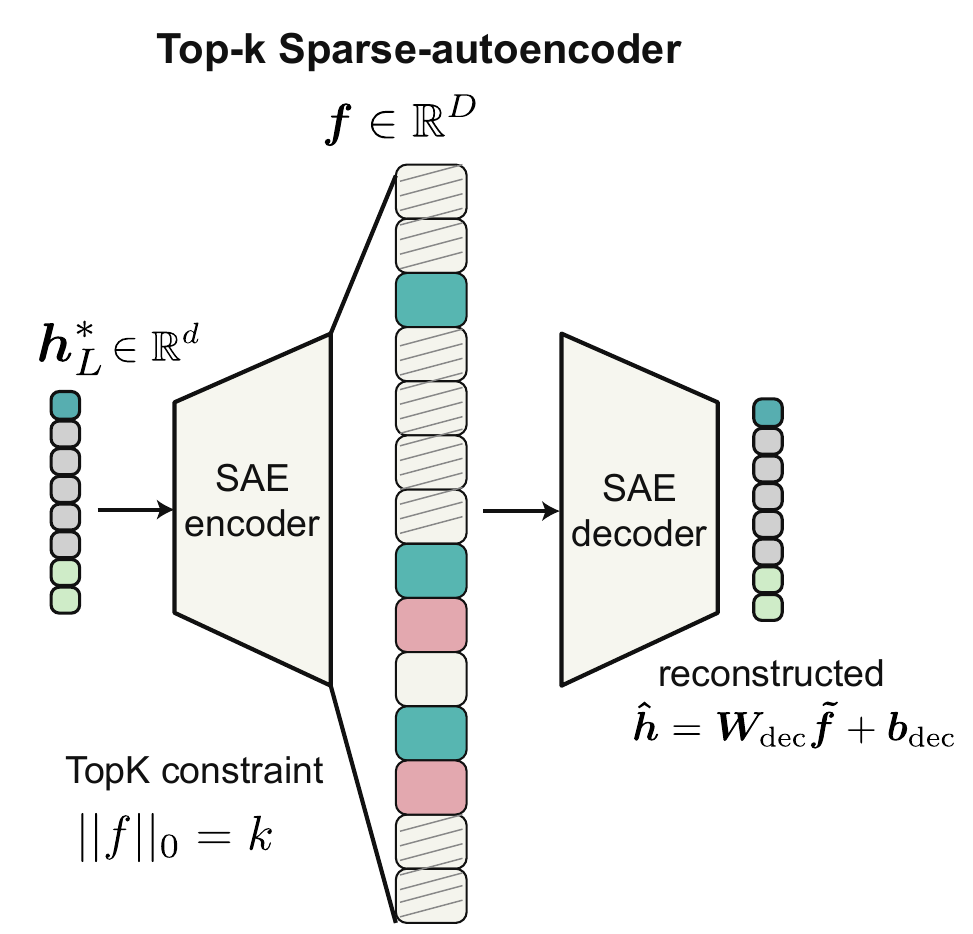}
\caption{TopK SAE training at $L^*$. The answer-position residual stream is encoded, sparsified by TopK, and decoded back to the residual-stream space. The trained feature basis is used for AtP$^*$ attribution and $S_{0.95}$ clamp interventions.}
\label{fig:sae-training}
\end{figure}

Figure~\ref{fig:sae-training} shows the training objective. For each model we train one expansion-8 TopK SAE ($k=64$) to minimize $\|h-\hat h(f(h))\|^2$ on per-token $L^*$ activations sampled from a multi-corpus pool. The pool comprises LibriSpeech \citep{panayotov2015librispeech}, IViE, JL-Corpus, CREMA-D, ESD-English, RAVDESS \citep{livingstone2018ravdess}, and VESUS. Each prosody corpus is capped at 300k training-split tokens and LibriSpeech at $\approx\!0.8$M ($\approx\!42\%$ of the pool) \citep{aparin2026audiosae}. LibriSpeech (read speech) and RAVDESS broaden the training pool only; they are not evaluated as mechanism cells. For every corpus, 30\% of clips are held out by content group (speaker $\times$ transcript; IViE by sentence pair). No test clip therefore shares an actor or sentence with training. The SAEs pass the canonical diagnostics: feature death $<5\%$ (0 dead), $L_0=k=64$, and pooled $R^2$ between 0.94 and 0.98. Crucially, held-out VESUS answer-position reconstruction reaches $R^2\approx0.83$--0.88. This is the coverage that earlier LibriSpeech-only SAEs lacked (negative held-out VESUS $R^2$), which motivated the multi-corpus retraining used throughout \S\ref{sec:sfc}.

\subsection{AtP$^*$ Attribution and Subspace Selection}
\label{app:sfc-attribution}
For each clip $i$, we insert the SAE reconstruction at the
answer-position state and compute the prosodic log-odds
\[
m_i =
\mathrm{logsumexp}(z^{(i)}_+)-
\mathrm{logsumexp}(z^{(i)}_-),
\]
where $z^{(i)}_+$ and $z^{(i)}_-$ are the logits of the positive- and negative-pole answer tokens. The quantity $m_i$ is then backpropagated to the SAE feature vector to compute the AtP$^*$ score:
\[
\mathrm{attr}_{i,j}
=
f^{(i)}_j
\cdot
\frac{\partial m_i}{\partial f^{(i)}_j}.
\]
Class-level attribution is then
\[
\mu^{\pm}_j =
\frac{1}{|C_{\pm}|}\sum_{i\in C_{\pm}}\mathrm{attr}_{i,j},
\qquad
\mathrm{net\_attr}_j=\mu^+_j-\mu^-_j .
\]
Features are sorted by descending $|\mathrm{net\_attr}|$. Here the set $S_{\tau}$ is the smallest prefix whose cumulative absolute attribution reaches fraction $\tau$ of the total. We use $S_{0.95}$ throughout.

\begin{figure}[t]
\centering
\includegraphics[width=\columnwidth]{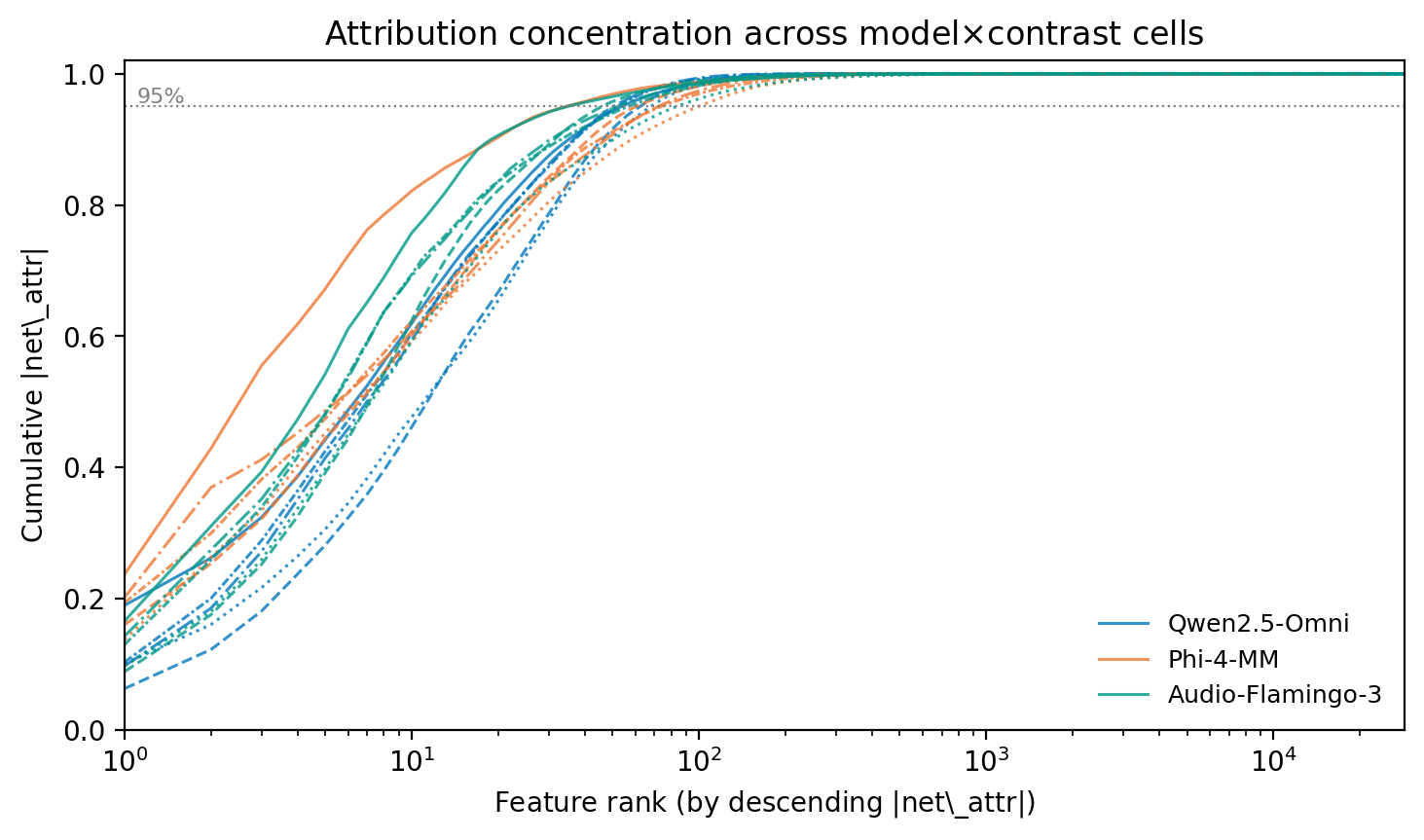}
\caption{Attribution concentration across model$\times$contrast cells under the multi-corpus SAE, including the held-out VESUS cells. Features are ordered by descending $|\mathrm{net\_attr}|$ and the curve is cumulative; the dotted line marks the 95\% threshold defining $S_{0.95}$. Qwen, Phi-4, and AF3 are shown (full per-feature attribution available); DeSTA's $|S_{0.95}|$ on its clean cells (52--58) is reported in Table~\ref{tab:sfc-descriptors-desta}. Every cell reaches 95\% within $\sim$100 features of dictionaries of 24{,}576--28{,}672 features.}
\label{fig:sfc-attribution}
\end{figure}

\subsection{Evaluation Split and Robustness Checks}
\label{app:sfc-evaluation}
For each cell, $S_{0.95}$ and the activation means $\mu^{+}_{\mathrm{act}},\mu^{-}_{\mathrm{act}}$ are estimated on the train portion only. The reported clamp results are evaluated on held out test clips. Full positive-class $\alpha$-sweeps per cell are in Table~\ref{tab:sfc-recovery}.

\begin{table*}[t]
\small
\centering
\setlength{\tabcolsep}{3pt}
\begin{tabular}{llrrrrrr}
\toprule
Model & Cell & $\alpha{=}0$ & $\alpha{=}1$ & $\alpha{=}2$ & $\alpha{=}3$ & $\alpha{=}5$ & $\alpha{=}8$ \\
\midrule
Qwen2.5-Omni & IViE Q/stmt & 0.465 & 0.746 & 0.817 & 0.930 & \textbf{1.000} & \textbf{1.000} \\
Qwen2.5-Omni & CREMA-D happy/sad & 0.872 & 0.997 & \textbf{1.000} & \textbf{1.000} & \textbf{1.000} & \textbf{1.000} \\
Qwen2.5-Omni & CREMA-D neutral/angry & 0.166 & 0.578 & \textbf{1.000} & \textbf{1.000} & \textbf{1.000} & \textbf{1.000} \\
Qwen2.5-Omni & VESUS happy/sad$^\dagger$ & 0.394 & 0.839 & 0.926 & 0.957 & \textbf{1.000} & \textbf{1.000} \\
Qwen2.5-Omni & VESUS neutral/angry$^\dagger$ & 0.122 & 0.230 & 0.289 & 0.393 & 0.983 & \textbf{1.000} \\
\midrule
Phi-4-MM & IViE Q/stmt & 0.014 & 0.028 & 0.268 & 0.563 & 0.972 & \textbf{1.000} \\
Phi-4-MM & CREMA-D happy/sad & 0.888 & 0.967 & \textbf{1.000} & \textbf{1.000} & \textbf{1.000} & \textbf{1.000} \\
Phi-4-MM & CREMA-D neutral/angry & 0.330 & 0.632 & 0.962 & \textbf{1.000} & \textbf{1.000} & \textbf{1.000} \\
Phi-4-MM & VESUS happy/sad$^\dagger$ & 0.957 & 0.970 & 0.975 & 0.980 & 0.987 & \textbf{0.993} \\
Phi-4-MM & VESUS neutral/angry$^\dagger$ & 0.088 & 0.109 & 0.136 & 0.164 & 0.311 & \textbf{0.775} \\
\midrule
Audio-Flamingo-3 & IViE Q/stmt & 0.634 & 0.775 & 0.887 & 0.972 & \textbf{1.000} & \textbf{1.000} \\
Audio-Flamingo-3 & CREMA-D happy/sad & 0.975 & \textbf{1.000} & \textbf{1.000} & \textbf{1.000} & \textbf{1.000} & \textbf{1.000} \\
Audio-Flamingo-3 & CREMA-D neutral/angry & 0.973 & \textbf{1.000} & \textbf{1.000} & \textbf{1.000} & \textbf{1.000} & \textbf{1.000} \\
Audio-Flamingo-3 & VESUS happy/sad$^\dagger$ & 0.679 & 0.984 & \textbf{1.000} & \textbf{1.000} & \textbf{1.000} & \textbf{1.000} \\
Audio-Flamingo-3 & VESUS neutral/angry$^\dagger$ & 0.568 & 0.808 & \textbf{1.000} & \textbf{1.000} & \textbf{1.000} & \textbf{1.000} \\
\midrule
DeSTA2.5 & IViE Q/stmt & 0.028 & \textbf{1.000} & \textbf{1.000} & \textbf{1.000} & \textbf{1.000} & \textbf{1.000} \\
DeSTA2.5 & CREMA-D h/s, n/a & \multicolumn{6}{c}{\emph{masked (in-training)}} \\
DeSTA2.5 & VESUS happy/sad$^\dagger$ & 0.552 & 0.823 & 0.957 & \textbf{1.000} & \textbf{1.000} & \textbf{1.000} \\
DeSTA2.5 & VESUS neutral/angry$^\dagger$ & 0.593 & 0.739 & 0.968 & \textbf{1.000} & \textbf{1.000} & \textbf{1.000} \\
\bottomrule
\end{tabular}
\caption{SAE-clamp recovery on the positive class, full $\alpha$-sweep under the $S_{0.95}$ additive-diff clamp ($f \mathrel{+}= \alpha(\mu^+-\mu^-)$), multi-corpus SAE, evaluated on held-out test clips under the five-prompt majority vote (\S\ref{sec:setup-prompts}). Bold marks the peak per cell. $\dagger$ = held-out VESUS. DeSTA $\times$ CREMA-D is masked as in-training. All cells vote over the full five-prompt bank except DeSTA$\times$IViE, which votes over four (the canonical prompt's per-clip log-odds were not retained for that cell).}
\label{tab:sfc-recovery}
\end{table*}

\subsection{Acoustic Descriptors}
\label{app:sfc-acoustic}
Six per-clip acoustic descriptors are computed from the audio waveform:
\begin{itemize}
\setlength{\itemsep}{2pt}
\item \texttt{f0\_slope}: the linear-regression slope of voiced-frame $F_0$ over frame index (Hz/s, a question-intonation diagnostic).
\item \texttt{f0\_term}: terminal $F_0$ shift, defined as mean $F_0$ of the last 25\% of voiced frames minus mean of the first 75\% (positive $\Rightarrow$ terminal rise).
\item \texttt{dur\_s}: clip duration in seconds.
\item \texttt{rms\_mean}: mean frame-level RMS energy.
\item \texttt{rms\_std}: standard deviation of frame-level RMS.
\item \texttt{rms\_max\_min}: RMS dynamic range, $\max-\min$.
\end{itemize}

\subsection{Per-Model Mechanism Grids}
\label{app:sfc-mechanism-grids}
\begin{figure*}[t]
\centering
\includegraphics[width=\textwidth]{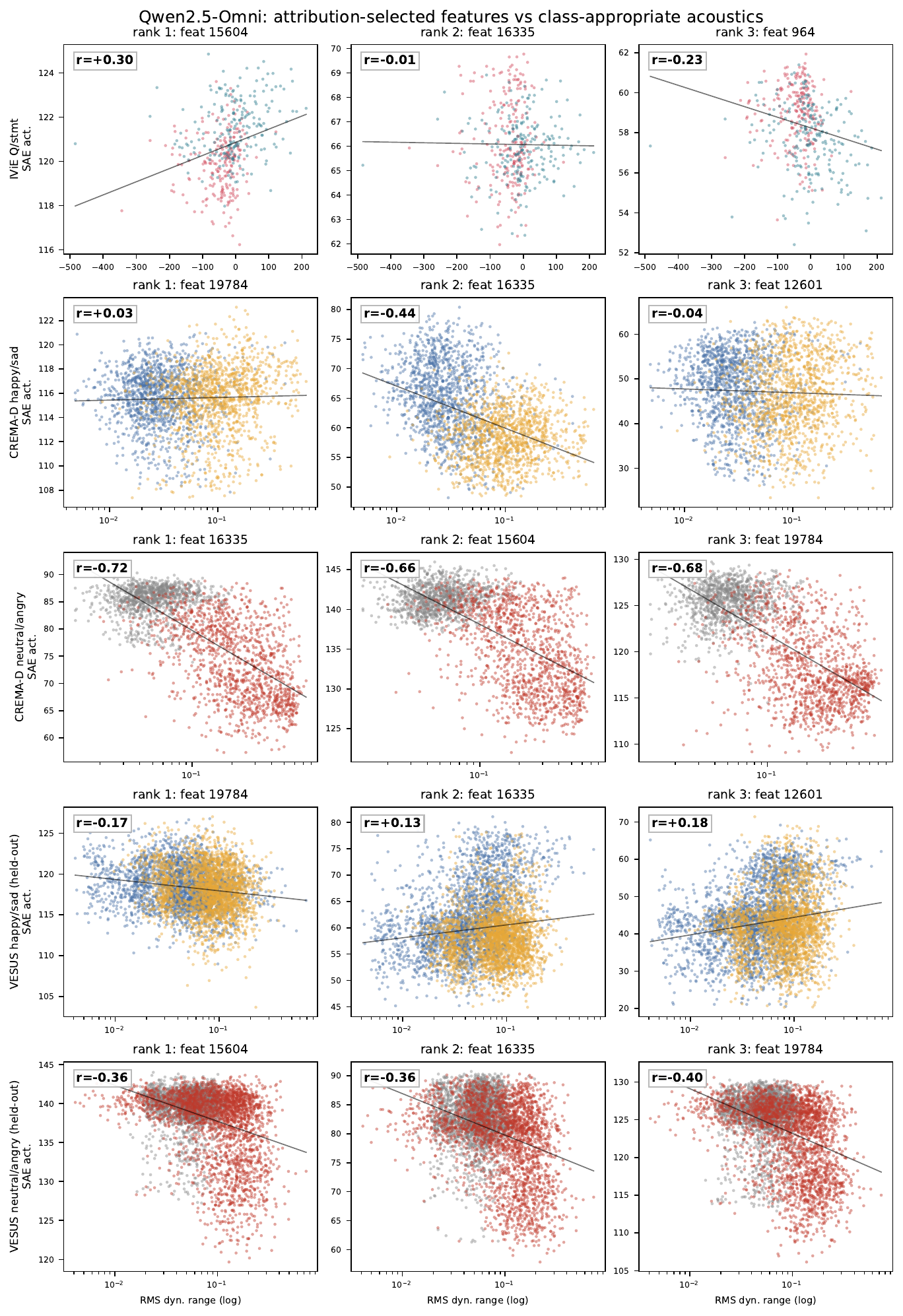}
\caption{Qwen2.5-Omni: top-3 attribution features per cell vs.\ class-appropriate acoustic descriptors (F0 slope for Q/stmt; RMS dynamic range for emotion).}
\label{fig:mech-grid-qwen}
\end{figure*}

\begin{figure*}[t]
\centering
\includegraphics[width=\textwidth]{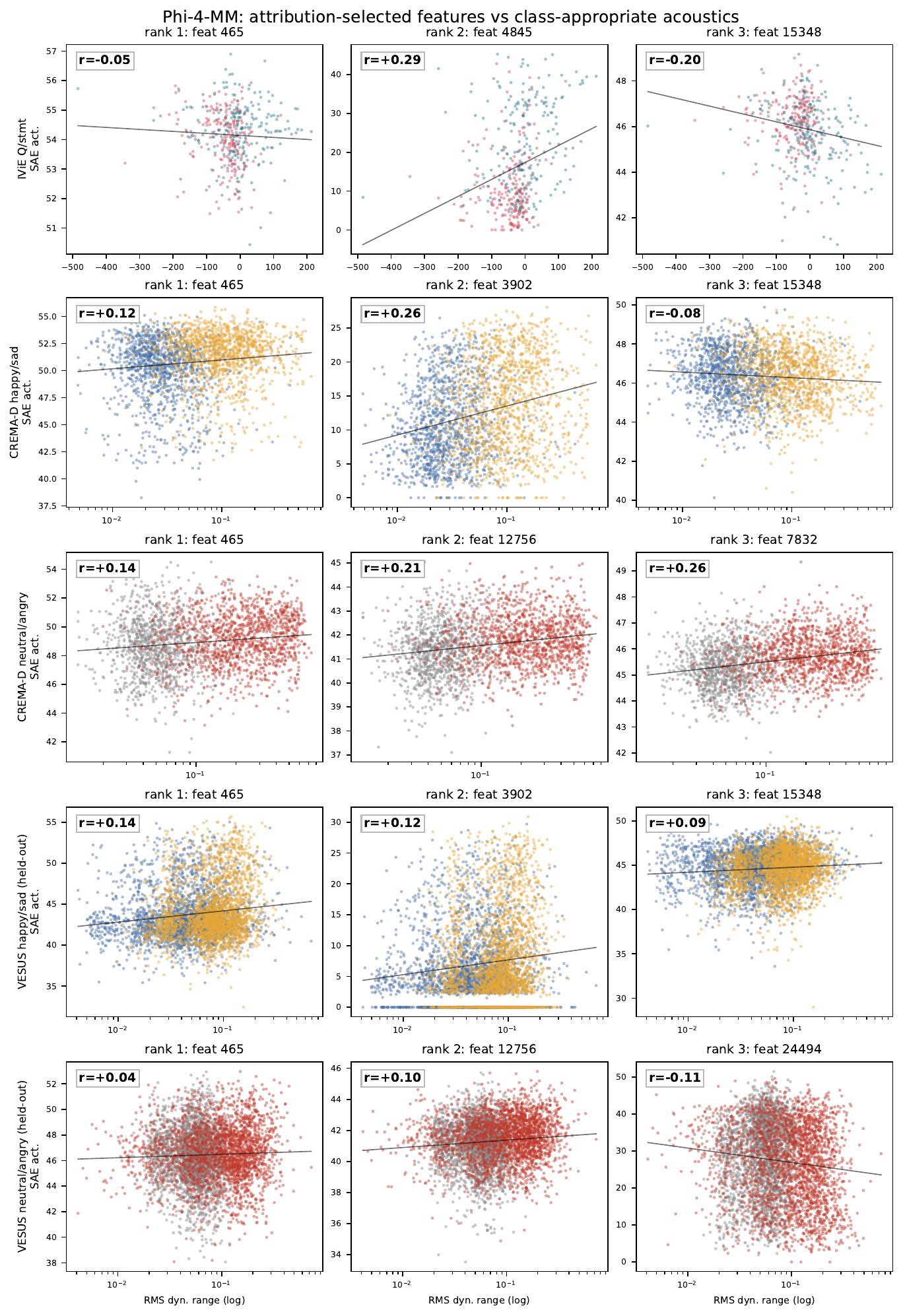}
\caption{Phi-4-MM: top-3 attribution features per cell. No feature reaches $|r|>0.5$ with the acoustic descriptors (cf.\ Table~\ref{tab:sfc-descriptors-phi4}).}
\label{fig:mech-grid-phi4}
\end{figure*}

\begin{figure*}[t]
\centering
\includegraphics[width=\textwidth]{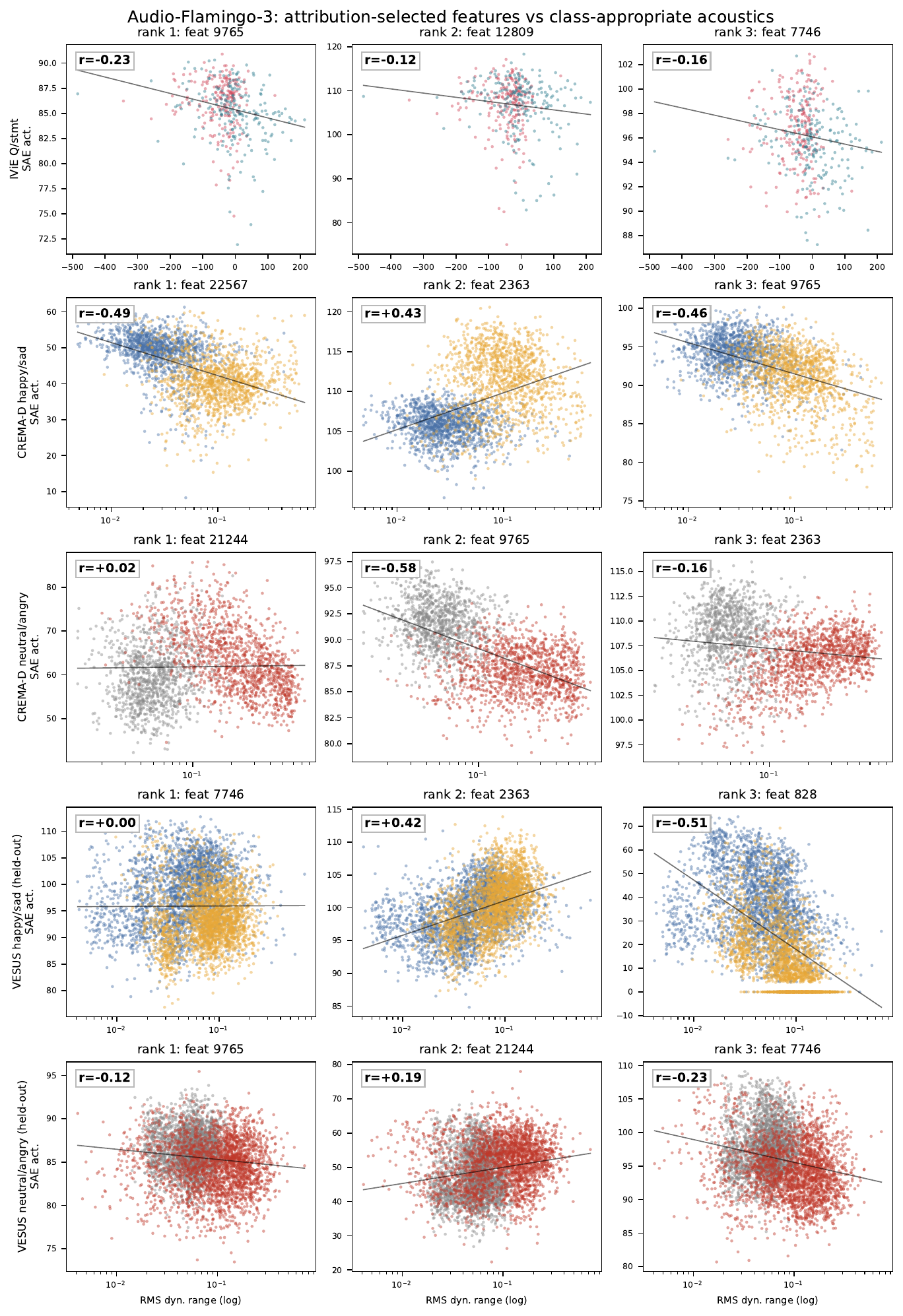}
\caption{Audio-Flamingo-3: top-3 attribution features per cell vs.\ class-appropriate acoustic descriptors.}
\label{fig:mech-grid-af3}
\end{figure*}

\begin{figure*}[t]
\centering
\includegraphics[width=\textwidth]{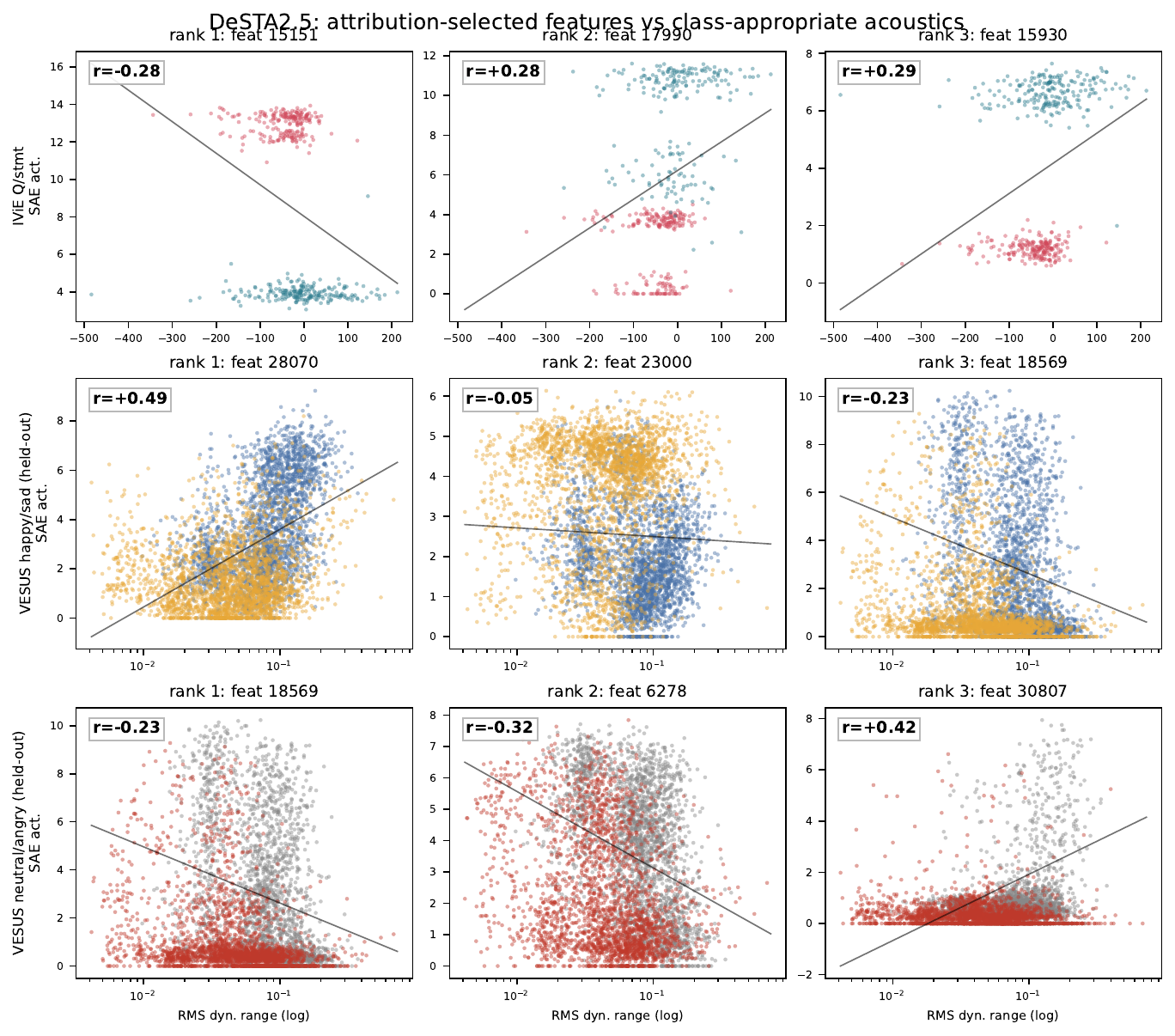}
\caption{DeSTA2.5: top-3 attribution features per cell vs.\ class-appropriate acoustic descriptors. CREMA-D cells are masked as in-training.}
\label{fig:mech-grid-desta}
\end{figure*}

Figures~\ref{fig:mech-grid-qwen}--\ref{fig:mech-grid-desta} show, for each model, the top-3 attribution features per cell, each plotted against the class-appropriate acoustic descriptor. Rows are contrasts (including the held-out VESUS cells); columns are attribution ranks. The main-text Figure~\ref{fig:sae-mechanism-grid} shows the single strongest acoustic-correlate panel per cell.

\subsection{Per-Cell Descriptor Correlations}
\label{app:sfc-desc}
Tables~\ref{tab:sfc-descriptors-qwen}--\ref{tab:sfc-descriptors-desta} report full-sample Pearson correlations between each model's top-3 attribution features per cell and the six acoustic descriptors. Correlations are computed under the multi-corpus SAE used in Figure~\ref{fig:sae-mechanism-grid}. Feature IDs and correlations therefore match the figure and the per-model grids in Appendix~\ref{app:sfc-mechanism-grids}. For the CREMA-D and VESUS RMS columns (\texttt{rms\_mean}, \texttt{rms\_std}, \texttt{rms\_max\_min}) the correlation is computed against $\log(\text{RMS})$. All other entries are raw-space Pearson. Bold entries mark $|r|>0.5$. Strong class-appropriate alignment is concentrated on the emotion cells (Qwen and AF3 on CREMA-D; AF3 on held-out VESUS happy/sad). DeSTA's VESUS correlations peak just below the threshold ($|r|=0.49$). The IViE cells are weak for every model ($|r|\le0.30$, with AF3's IViE features tracking duration rather than F0), and Phi-4 has no entry above $|r|=0.29$. As elsewhere, DeSTA's CREMA-D cells are masked as in-training.

\input{tab_sfc_descriptors_4models}

%% file: tab_audio_path_numbers.tex
\begin{table}[t]
\centering\scriptsize\setlength{\tabcolsep}{3.5pt}
\begin{tabular}{llccccc}
\toprule
Model & Cell & last & peak & @ & PROJ & lift \\
\midrule
Qwen2.5-Omni & IViE Q/stmt & 0.78 & 0.79 & 0.84 & 0.81 & +0.26 \\
 & ESD & 0.79 & 0.81 & 0.97 & 0.84 & +0.35 \\
 & VESUS & 0.73 & 0.73 & 0.72 & 0.71 & +0.29 \\
 & CREMA-D & 0.85 & 0.85 & 1.00 & 0.88 & +0.28 \\
\midrule
Audio-Flamingo-3 & IViE Q/stmt & 0.65 & 0.71 & 0.97 & 0.71 & +0.19 \\
 & ESD & 0.79 & 0.81 & 0.91 & 0.79 & +0.29 \\
 & VESUS & 0.69 & 0.71 & 0.97 & 0.69 & +0.31 \\
 & CREMA-D & 0.92 & 0.92 & 1.00 & 0.92 & +0.35 \\
\midrule
DeSTA2.5-Audio & IViE Q/stmt & 0.61 & 0.79 & 0.81 & 0.76 & +0.25 \\
 & ESD$^\dagger$ & \multicolumn{5}{c}{\textit{in-training data (masked)}} \\
 & VESUS & 0.71 & 0.73 & 0.91 & 0.72 & +0.32 \\
 & CREMA-D$^\dagger$ & \multicolumn{5}{c}{\textit{in-training data (masked)}} \\
\midrule
Phi-4-MM & IViE Q/stmt & 0.78 & 0.81 & 0.75 & 0.79 & +0.21 \\
 & ESD & 0.63 & 0.82 & 0.50 & 0.64 & +0.15 \\
 & VESUS & 0.38 & 0.52 & 0.00 & 0.42 & -0.01 \\
 & CREMA-D & 0.76 & 0.84 & 0.62 & 0.54 & -0.06 \\
\bottomrule
\end{tabular}
\caption{Audio-path probe WA per (model, cell), canonical 4-class emotion (CREMA-D/ESD/VESUS) and 2-class IViE. \emph{last} = final encoder layer, \emph{peak} = best layer at normalized depth \emph{@}, \emph{PROJ} = projector output passed to the LLM, \emph{lift} = PROJ trained $-$ random-init. $^\dagger$ \emph{masked}: the corpus is in DeSTA2.5's training data, so the cell is excluded from cross-model claims.}
\label{tab:audio-path-numbers}
\end{table}

%% file: tab_di_slope_ci_4model.tex
\begin{table}[t]
\scriptsize\centering\setlength{\tabcolsep}{3pt}
\begin{tabular}{llc}
\toprule
Model & Contrast & Slope [95\% CI] \\
\midrule
Qwen & IViE Q/stmt & $+1.34\,[+1.23,\,+1.45]$ \\
Qwen & CREMA-D h/s & $+1.76\,[+1.07,\,+2.44]$ \\
Qwen & CREMA-D n/a & $+2.01\,[+1.31,\,+2.71]$ \\
Qwen & VESUS h/s & $+0.95\,[+0.87,\,+1.03]$ \\
Qwen & VESUS n/a & $+1.09\,[+1.03,\,+1.16]$ \\
\midrule
Phi-4 & IViE Q/stmt & $+1.84\,[+1.28,\,+2.41]$ \\
Phi-4 & CREMA-D h/s & $+0.64\,[+0.52,\,+0.75]$ \\
Phi-4 & CREMA-D n/a & $+0.48\,[+0.40,\,+0.57]$ \\
Phi-4 & VESUS h/s & $+0.47\,[+0.46,\,+0.47]$ \\
Phi-4 & VESUS n/a & $+0.31\,[+0.24,\,+0.39]$ \\
\midrule
AF3 & IViE Q/stmt & $+2.26\,[+1.71,\,+2.81]$ \\
AF3 & CREMA-D h/s & $+3.20\,[+1.79,\,+4.61]$ \\
AF3 & CREMA-D n/a & $+2.57\,[+0.50,\,+4.64]$ \\
AF3 & VESUS h/s & $+2.04\,[+1.48,\,+2.59]$ \\
AF3 & VESUS n/a & $+2.80\,[+1.29,\,+4.30]$ \\
\midrule
DeSTA & IViE Q/stmt & $+0.29\,[+0.28,\,+0.30]$ \\
DeSTA & CREMA-D h/s & \emph{masked (in-training)} \\
DeSTA & CREMA-D n/a & \emph{masked (in-training)} \\
DeSTA & VESUS h/s & $+1.83\,[+1.81,\,+1.85]$ \\
DeSTA & VESUS n/a & $+2.92\,[+2.90,\,+2.93]$ \\
\bottomrule
\end{tabular}
\caption{Direction-injection slope per cell with 95\% confidence intervals. The CI is a $t$-based OLS interval for the slope of the mean answer-token log-odds across the sampled $\alpha$ grid. All intervals exclude zero.}
\label{tab:di-slope-ci}
\end{table}

%% file: tab_sfc_descriptors_4models.tex
\begin{table*}[t]
\centering
\scriptsize
\setlength{\tabcolsep}{3pt}
\begin{tabular}{llrrrrrrr}
\toprule
Cell & feat & $|\text{attr}|$ & f0\_slope & f0\_term & dur\_s & rms\_mean & rms\_std & rms\_max\_min \\
\midrule
IViE Q/stmt ($|S_{0.95}|=59$) & 15604 & 0.44 & +0.30 & +0.29 & -0.20 & -0.05 & -0.07 & -0.09 \\
 & 16335 & 0.17 & -0.01 & -0.01 & -0.07 & +0.07 & +0.06 & +0.06 \\
 & 964 & 0.15 & -0.23 & -0.24 & +0.12 & -0.00 & +0.02 & +0.03 \\
\midrule
CREMA-D happy/sad ($|S_{0.95}|=62$) & 19784 & 0.44 & +0.11 & +0.10 & -0.11 & +0.02 & +0.03 & +0.03 \\
 & 16335 & 0.42 & -0.10 & -0.07 & +0.16 & -0.44 & -0.47 & -0.44 \\
 & 12601 & 0.42 & +0.00 & -0.02 & -0.12 & -0.03 & -0.05 & -0.04 \\
\midrule
CREMA-D neutral/angry ($|S_{0.95}|=51$) & 16335 & 0.80 & -0.11 & -0.13 & -0.04 & \textbf{-0.72} & \textbf{-0.73} & \textbf{-0.72} \\
 & 15604 & 0.73 & -0.09 & -0.10 & -0.08 & \textbf{-0.67} & \textbf{-0.67} & \textbf{-0.66} \\
 & 19784 & 0.72 & -0.05 & -0.09 & -0.11 & \textbf{-0.67} & \textbf{-0.69} & \textbf{-0.68} \\
\midrule
VESUS happy/sad ($|S_{0.95}|=69$) & 19784 & 0.24 & +0.04 & +0.11 & -0.30 & -0.19 & -0.14 & -0.17 \\
 & 16335 & 0.14 & -0.02 & -0.03 & +0.07 & +0.10 & +0.12 & +0.13 \\
 & 12601 & 0.13 & -0.04 & -0.02 & -0.01 & +0.15 & +0.17 & +0.18 \\
\midrule
VESUS neutral/angry ($|S_{0.95}|=53$) & 15604 & 0.33 & +0.19 & +0.20 & +0.02 & -0.31 & -0.35 & -0.36 \\
 & 16335 & 0.31 & +0.19 & +0.19 & +0.10 & -0.29 & -0.35 & -0.36 \\
 & 19784 & 0.28 & +0.20 & +0.25 & -0.07 & -0.35 & -0.39 & -0.40 \\
\bottomrule
\end{tabular}
\caption{Qwen2.5-Omni: per-cell descriptor correlations for the top-3 attribution features (multi-corpus SAE; CREMA-D/VESUS RMS columns vs $\log$(RMS), others raw). Bold entries mark $|r|>0.5$.}
\label{tab:sfc-descriptors-qwen}
\end{table*}

\begin{table*}[t]
\centering
\scriptsize
\setlength{\tabcolsep}{3pt}
\begin{tabular}{llrrrrrrr}
\toprule
Cell & feat & $|\text{attr}|$ & f0\_slope & f0\_term & dur\_s & rms\_mean & rms\_std & rms\_max\_min \\
\midrule
IViE Q/stmt ($|S_{0.95}|=33$) & 465 & 1.92 & -0.05 & -0.03 & -0.09 & +0.06 & +0.08 & +0.07 \\
 & 4845 & 1.56 & +0.29 & +0.27 & +0.04 & +0.03 & +0.04 & +0.03 \\
 & 15348 & 1.03 & -0.20 & -0.24 & +0.18 & +0.11 & +0.09 & +0.10 \\
\midrule
CREMA-D happy/sad ($|S_{0.95}|=59$) & 465 & 0.59 & +0.10 & +0.10 & -0.06 & +0.11 & +0.13 & +0.12 \\
 & 3902 & 0.34 & +0.12 & +0.14 & -0.10 & +0.29 & +0.29 & +0.26 \\
 & 15348 & 0.25 & -0.00 & +0.00 & -0.05 & -0.10 & -0.09 & -0.08 \\
\midrule
CREMA-D neutral/angry ($|S_{0.95}|=69$) & 465 & 0.64 & +0.05 & +0.04 & -0.13 & +0.15 & +0.15 & +0.14 \\
 & 12756 & 0.54 & +0.03 & +0.05 & -0.16 & +0.22 & +0.21 & +0.21 \\
 & 7832 & 0.14 & +0.08 & +0.06 & -0.02 & +0.26 & +0.25 & +0.26 \\
\midrule
VESUS happy/sad ($|S_{0.95}|=102$) & 465 & 0.19 & -0.03 & -0.09 & +0.25 & +0.14 & +0.12 & +0.14 \\
 & 3902 & 0.17 & -0.00 & -0.07 & +0.29 & +0.11 & +0.10 & +0.12 \\
 & 15348 & 0.10 & -0.01 & -0.05 & +0.27 & +0.17 & +0.07 & +0.09 \\
\midrule
VESUS neutral/angry ($|S_{0.95}|=75$) & 465 & 0.43 & +0.00 & -0.03 & +0.02 & +0.01 & +0.04 & +0.04 \\
 & 12756 & 0.23 & -0.06 & -0.08 & +0.02 & +0.13 & +0.10 & +0.10 \\
 & 24494 & 0.18 & +0.03 & +0.14 & -0.33 & -0.08 & -0.09 & -0.11 \\
\bottomrule
\end{tabular}
\caption{Phi-4-MM: per-cell descriptor correlations for the top-3 attribution features (multi-corpus SAE; CREMA-D/VESUS RMS columns vs $\log$(RMS), others raw). Bold entries mark $|r|>0.5$.}
\label{tab:sfc-descriptors-phi4}
\end{table*}

\begin{table*}[t]
\centering
\scriptsize
\setlength{\tabcolsep}{3pt}
\begin{tabular}{llrrrrrrr}
\toprule
Cell & feat & $|\text{attr}|$ & f0\_slope & f0\_term & dur\_s & rms\_mean & rms\_std & rms\_max\_min \\
\midrule
IViE Q/stmt ($|S_{0.95}|=37$) & 9765 & 0.80 & -0.23 & -0.22 & +0.25 & +0.02 & +0.03 & +0.04 \\
 & 12809 & 0.70 & -0.12 & -0.15 & +0.37 & +0.05 & +0.06 & +0.05 \\
 & 7746 & 0.40 & -0.16 & -0.16 & +0.12 & +0.09 & +0.08 & +0.09 \\
\midrule
CREMA-D happy/sad ($|S_{0.95}|=47$) & 22567 & 1.80 & -0.12 & -0.07 & +0.24 & -0.48 & \textbf{-0.51} & -0.49 \\
 & 2363 & 1.77 & +0.21 & +0.21 & -0.30 & +0.46 & +0.46 & +0.43 \\
 & 9765 & 1.57 & -0.07 & -0.05 & +0.06 & -0.46 & -0.46 & -0.46 \\
\midrule
CREMA-D neutral/angry ($|S_{0.95}|=52$) & 21244 & 2.83 & -0.02 & +0.04 & +0.03 & -0.02 & +0.01 & +0.02 \\
 & 9765 & 2.61 & -0.09 & -0.14 & -0.19 & \textbf{-0.55} & \textbf{-0.58} & \textbf{-0.58} \\
 & 2363 & 1.55 & -0.03 & -0.08 & -0.22 & -0.10 & -0.14 & -0.16 \\
\midrule
VESUS happy/sad ($|S_{0.95}|=84$) & 7746 & 0.93 & +0.03 & +0.04 & -0.05 & -0.04 & +0.01 & +0.00 \\
 & 2363 & 0.80 & -0.09 & -0.08 & -0.12 & +0.40 & +0.43 & +0.42 \\
 & 828 & 0.77 & +0.11 & +0.09 & +0.16 & -0.45 & \textbf{-0.51} & \textbf{-0.51} \\
\midrule
VESUS neutral/angry ($|S_{0.95}|=58$) & 9765 & 1.80 & +0.02 & +0.05 & -0.05 & -0.04 & -0.10 & -0.12 \\
 & 21244 & 1.80 & -0.03 & -0.04 & -0.10 & +0.07 & +0.19 & +0.19 \\
 & 7746 & 1.11 & +0.12 & +0.19 & -0.34 & -0.24 & -0.20 & -0.23 \\
\bottomrule
\end{tabular}
\caption{Audio-Flamingo-3: per-cell descriptor correlations for the top-3 attribution features (multi-corpus SAE; CREMA-D/VESUS RMS columns vs $\log$(RMS), others raw). Bold entries mark $|r|>0.5$.}
\label{tab:sfc-descriptors-af3}
\end{table*}

\begin{table*}[t]
\centering
\scriptsize
\setlength{\tabcolsep}{3pt}
\begin{tabular}{llrrrrrrr}
\toprule
Cell & feat & $|\text{attr}|$ & f0\_slope & f0\_term & dur\_s & rms\_mean & rms\_std & rms\_max\_min \\
\midrule
IViE Q/stmt ($|S_{0.95}|=54$) & 15151 & 8.79 & -0.28 & -0.24 & -0.02 & -0.03 & -0.05 & -0.04 \\
 & 17990 & 6.14 & +0.28 & +0.28 & -0.11 & +0.01 & +0.04 & +0.03 \\
 & 15930 & 5.11 & +0.29 & +0.26 & +0.01 & +0.04 & +0.06 & +0.04 \\
\midrule
CREMA-D happy/sad, neutral/angry & \multicolumn{8}{c}{\emph{masked (in-training)}} \\
\midrule
VESUS happy/sad ($|S_{0.95}|=58$) & 28070 & 2.10 & -0.14 & -0.23 & +0.13 & +0.42 & +0.47 & +0.49 \\
 & 23000 & 1.64 & +0.02 & -0.07 & +0.32 & -0.08 & -0.07 & -0.05 \\
 & 18569 & 1.63 & +0.08 & +0.13 & -0.09 & -0.20 & -0.22 & -0.23 \\
\midrule
VESUS neutral/angry ($|S_{0.95}|=52$) & 18569 & 3.18 & +0.08 & +0.13 & -0.09 & -0.20 & -0.22 & -0.23 \\
 & 6278 & 2.82 & +0.10 & +0.19 & -0.21 & -0.28 & -0.30 & -0.32 \\
 & 30807 & 2.81 & -0.12 & -0.24 & +0.16 & +0.35 & +0.40 & +0.42 \\
\bottomrule
\end{tabular}
\caption{DeSTA2.5: per-cell descriptor correlations for the top-3 attribution features (multi-corpus SAE; VESUS RMS columns vs $\log$(RMS), others raw). Bold entries mark $|r|>0.5$. CREMA-D is masked as in-training.}
\label{tab:sfc-descriptors-desta}
\end{table*}